\documentclass[lettersize,journal]{IEEEtran}
\usepackage{amsmath,amsfonts}
\usepackage{amssymb}
\usepackage{amsthm}
\usepackage{dsfont}
\usepackage{array}
\usepackage[caption=false,font=normalsize,labelfont=sf,textfont=sf]{subfig}
\usepackage{stfloats}
\usepackage{url}
\usepackage{verbatim}
\usepackage{graphicx}
\usepackage{cite}
\usepackage[hidelinks]{hyperref}

\usepackage[caption=false]{subfig}
\usepackage{color}
\usepackage{dsfont}
\usepackage{booktabs}
\usepackage{bm}
\usepackage{multirow}
\usepackage{multicol}
\usepackage[abs]{overpic}
\usepackage{textcomp}
\usepackage{pifont}
\usepackage[table]{xcolor}
\usepackage{pifont}

\usepackage[linesnumbered,ruled,vlined]{algorithm2e}
\usepackage{algpseudocode}

\def\eg{\emph{e.g.,}}
\def\etal{\emph{et al.}}
\def\ie{\emph{i.e.,}}

\def\wrt{\emph{w.r.t.}}

\begin{document}

\title{Prototype Perturbation for Relaxing Alignment Constraints in Backward-Compatible Learning}

\author{Zikun~Zhou,
        Yushuai~Sun,
        Wenjie~Pei,~\IEEEmembership{Senior Member, IEEE,}
        Xin~Li,
        and Yaowei~Wang,~\IEEEmembership{Member, IEEE}
\thanks{
    This work was supported in part by the Guangdong Basic and Applied Basic Research Foundation (Grants No. 2025A1515010705 and 2024A1515011292), in part by the National Natural Science Foundation of China (Grants No. 62372133 and 62476148), and in part by the ``Guangdong Special Support Plan" (Grant No. 2023TQ07A784).

    Zikun Zhou, Yushuai Sun, Wenjie Pei, and Yaowei Wang are with Harbin Institute of Technology, Shenzhen, China. (zhouzikunhit@gamil.com; yushuai112233@gamil.com; wenjiecoder@outlook.com; wangyw@pcl.ac.cn). Zikun Zhou and Yaowei Wang are also with Pengcheng Laboratory, Shenzhen, China.
    Xin Li is with Pengcheng Laboratory, Shenzhen, China (xinlihitsz@gmail.com). 

    Zikun Zhou and Yushuai Sun contribute equally to this work. Wenjie Pei and Yaowei Wang are the corresponding authors.
    }
}


\maketitle

\begin{abstract}
The traditional paradigm to update retrieval models requires re-computing the embeddings of the gallery data, a time-consuming and computationally intensive process known as backfilling. To circumvent backfilling, Backward-Compatible Learning (BCL) has been widely explored, which aims to train a new model compatible with the old one. Many previous works focus on effectively aligning the embeddings of the new model with those of the old one to enhance backward compatibility. Nevertheless, such strong alignment constraints would compromise the discriminative ability of the new model, particularly when different classes are closely clustered and hard to distinguish in the old feature space. To address this issue, we propose to relax the constraints by introducing perturbations to the old feature prototypes. This allows us to align the new feature space with a pseudo-old feature space defined by these perturbed prototypes, thereby preserving the discriminative ability of the new model in backward-compatible learning. We have developed two approaches for calculating the perturbations: Neighbor-Driven Prototype Perturbation (NDPP) and Optimization-Driven Prototype Perturbation (ODPP). Particularly, they take into account the feature distributions of not only the old but also the new models to obtain proper perturbations along with new model updating. Extensive experiments on the landmark and commodity datasets demonstrate that our approaches perform favorably against state-of-the-art BCL algorithms. The code is available at \url{https://github.com/Bunny-Black/Prototype-Perturbation}.
\end{abstract}


\section{Introduction}
\begin{figure}[t!]
\centering
\includegraphics[width=0.99\linewidth]{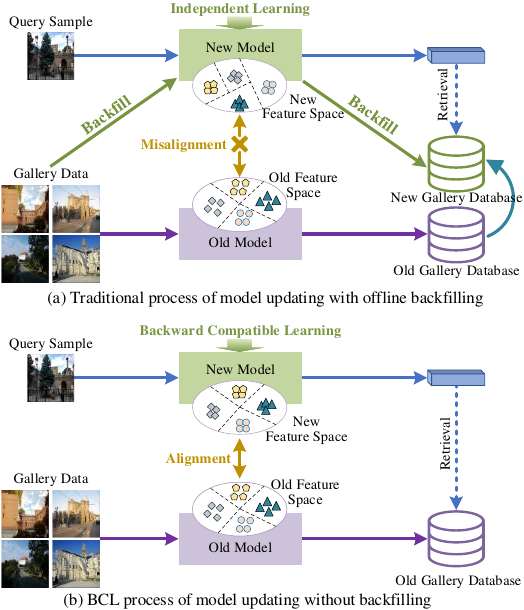}
\vspace{-5mm}
\caption{Two different paradigms for updating the embedding model for the retrieval system. (a) The new model is trained independently and used to re-extract the embeddings of all gallery data,~\ie~backfilling, which is time-consuming and computationally intensive. (b) With backward-compatible learning, the query embedding is directly comparable with the existing gallery embeddings without backfilling.}
\label{Fig:introduction}
\vspace{-3mm}
\end{figure}

Image retrieval has drawn much attention due to its wide range of applications~\cite{xu20213rd,yuqi20212nd,extending,sun2025learning}, such as E-commerce search and landmark localization. The retrieval systems typically employ an embedding model to transform raw data into high-dimensional vectors and maintain a large-scale gallery database storing the embeddings of the gallery data. Upon receiving a query sample, the online server calculates its embedding vector and compares the embedding with those in the gallery database to retrieve the most similar entries. As new training data or advanced model designs become available, retrieval service providers often desire to train a new model for better performance. However, if the new model is trained independently, its feature space naturally misaligns with that of the old one~\cite{advbct}, being incompatible with existing gallery embeddings. Hence, the system needs to re-extract the embeddings of the gallery data with the new model, a process named ``backfilling'', as shown in Figure~\ref{Fig:introduction}~(a). Nevertheless, backfilling is time-consuming and computationally intensive for large-scale galleries. 

\begin{figure}[t]
\centering
\includegraphics[width=1.0\columnwidth]{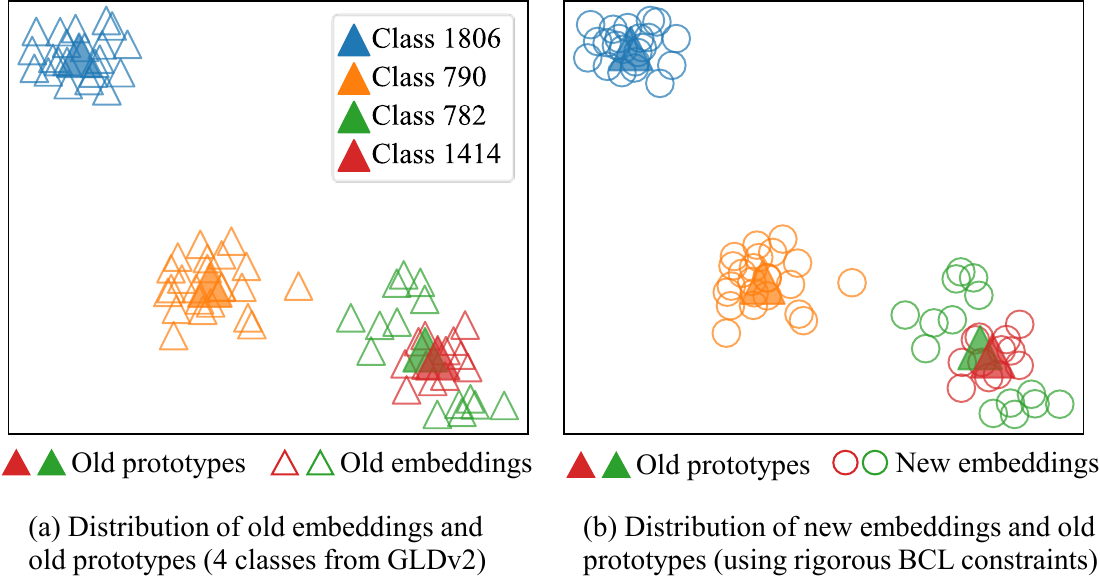}
\caption{Distribution of the embeddings and prototypes of 4 classes from GLDv2~\cite{gldv2}, which is visualized with t-SNE~\cite{t-SNE}. (a) Classes 782 \& 1414 are nearly indistinguishable in the old feature space. (b) The new model is trained with rigorous alignment constraints to the old feature space, producing new embeddings strictly surrounding the old prototypes. As a result, the new embeddings of these classes are clustered closely and hard to distinguish.}
\label{Fig:feature distribution}
\vspace{-4mm}
\end{figure}

To address this problem, Shen~\etal~\cite{BCT} propose the Backward-Compatible Learning (BCL) methodology; it enables the new model to produce embeddings comparable with existing gallery embeddings, allowing for backfilling-free model updating, as shown in Figure~\ref{Fig:introduction}~(b). To be specific, Shen~\etal~\cite{BCT} use the classifier of the old model to regularize the new model training to achieve backward compatibility. Subsequent studies further explore more sophisticated BCL methods~\cite{LCE, unibct,hot-refresh}. For example, AdvBCT~\cite{advbct} integrates adversarial learning into new model training, coupled with an elastic boundary mechanism to constrain the distribution of new embeddings. Instead of regularizing a new model with the old one, recent studies~\cite{biondi2024stationary,Cores,CL2R} opt to pre-allocate a fixed classifier based on regular polytopes to learn stationary features to achieve backward compatibility, which exhibit outstanding performance in sequential model updates.

Many aforementioned methods essentially enforce the embeddings of the new model to strictly align with those of the old model to enhance backward compatibility. Despite the impressive performance, they pay less attention to the potential adverse effect of the old feature distribution on the new model during BCL, which may hurt the discriminative ability of the new model. For instance, if two different classes are distributed closely and become nearly indistinguishable in the old feature space, imposing strict alignment constraints between the new and old models would cause these classes to remain indistinguishable in the new feature space. Figure~\ref{Fig:feature distribution} (a) and (b) illustrate such a case, where the new embeddings of Classes 782 \& 1414 become indistinguishable due to the overstrict alignment constraints.

To alleviate the adverse effect of the indistinguishable classes in the old feature space, we propose introducing perturbations to the old feature space while training the new model, enabling backward-compatible learning with a more flexible alignment constraint. Given that individual sample features may be noisy in representing the feature distribution, we opt to apply perturbations to the old class centers, \ie~old prototypes, which are robust to outliers. Technically, we impose perturbations on old prototypes to adaptively push them away from their indistinguishable neighbors, as shown in Figure~\ref{Fig:illustration_NDPP} (a). Then, we can constrain the new feature space to align with the pseudo-old feature space characterized by the perturbed old prototypes, learning a new model with strong discriminative ability. The crux of this idea lies in determining appropriate directions and amplitudes for the perturbations to enhance the discriminative ability of the new model without compromising its backward compatibility.

In this paper, we devise two implementations for prototype perturbation: Neighbor-Driven Prototype Perturbation (NDPP) and Optimization-Driven Prototype Perturbation (ODPP), which both calculate perturbations based on the similarities between prototypes. NDPP employs a heuristic approach to calculate the perturbation for an old prototype based on the repulsion from its neighboring prototypes, while ODPP learns the perturbations for old prototypes by optimizing an objective function designed to push similar prototypes away from each other. Notably, NDPP and ODPP leverage the prototypes of not only the old model but also the new model to determine the perturbations during BCL. This design allows for continual adjustment of perturbations conditioned on the new feature space, producing more suitable perturbations to enhance the discriminative ability of the new model. We conduct extensive experiments on diverse benchmarks~\cite{gldv2, rparis&roxford,inshop,Market1501,RSTPReid}. The experimental results demonstrate that NDPP and ODPP both substantially improve the retrieval performance of the new model without compromising the cross-model retrieval performance.
Our contributions are summarized as follows:
\begin{itemize}
    \item We propose a prototype perturbation mechanism to adaptively relax the alignment constraints for BCL to enhance the discriminative ability of the new model.
    \item We develop two novel approaches, neighbor-driven prototype perturbation and optimization-driven prototype perturbation, to implement the prototype perturbation mechanism; they leverage both the old and new prototypes to produce effective perturbations for robust BCL.
    \item Extensive experiments on diverse datasets demonstrate that both NDPP and ODPP perform favorably against state-of-the-art BCL methods, which showcases the effectiveness of our prototype perturbation mechanism.
\end{itemize}

\section{Related Work}
\noindent\textbf{Backward-compatible learning.}
To avoid backfilling for retrieval model updating, Shen~\etal~\cite{BCT} propose the backward-compatible learning method to train a new embedding model compatible with the old one. It regularizes the learning of the new model with the classifier of the old model. Afterward, many algorithms~\cite{LCE, DualTuning, hot-refresh, bt2, mixbct, jang2024towards} have been proposed to improve BCL performance. Several of them~\cite{DualTuning,hot-refresh,NCCL} resort to contrastive learning~\cite{oord2018representation} to achieve backward compatibility; they leverage the contrastive loss between the old models and new models as a regularization to acquire compatible representations. Besides contrastive learning, other sophisticated backward-compatible constraints have also been explored. For instance, AdvBCT~\cite{advbct} employs an elastic boundary to constrain the new embeddings and introduces adversarial learning to minimize the distribution disparity between new and old embeddings. RBCL~\cite{RBCL} proposes a ranking-based BCL method that directly optimizes the ranking metric between new and old features. SSPL~\cite{SSPL} designs a structure similarity preserving method to achieve feature compatibility between the query and gallery models. The above methods~\cite{BCT, LCE, DualTuning, NCCL,unibct} do not explicitly handle the indistinguishable classes in the old feature space, which would hurt the discriminative ability of the new model. To overcome the dilemma between backward compatibility and the new model performance, $\text{BT}^2$~\cite{bt2} opts to extend the representation of the new model with extra dimensions and proposes a basis transformation for BCL.

Most above methods typically regularize the training of a new model with a backward constraint based on the old model. Unlike them, recent studies~\cite{biondi2024stationary,Cores,CL2R} pre-allocate a fixed classifier based on regular polytopes to learn stationary features to achieve backward compatibility. The rationale is that stationarity ensures the consistency of feature distributions over time, enabling direct comparison between the embeddings from the old and updated models. Such methods demonstrate exceptional performance in scenarios involving continuous model updates. Furthermore, these stationary representation methods can be applied to the compatible lifelong learning problem, obtaining compatible features across different learning sessions within the lifelong learning paradigm~\cite{CL2R,biondi2024stationary}.

Notably, many studies in BCL are closely related to the concept of prototype, \ie~class center. Specifically, Dual-Tuning~\cite{DualTuning} designs a prototype-based contrastive loss to align the new embeddings with the old prototypes. SSPL~\cite{SSPL} leverages the centroid vectors of a product quantizer as anchor points to preserve structure similarity across models in the embedding space. To alleviate the effects of outlier samples on prototype calculation, UniBCT~\cite{unibct} designs a structural prototype refinement method that propagates the knowledge within neighboring intra-class samples, acquiring more accurate prototypes for BCL. The stationary representation methods~\cite{biondi2024stationary,Cores,CL2R} leverage a fixed classifier, \ie~fixed class prototypes, to pre-allocate the feature space to maintain the consistency of feature distributions over time. Unlike these methods leveraging static prototypes from the old model or the pre-allocated fixed classifier, our method continuously perturbs the prototypes of different classes during training, aiming to dynamically construct a more distinguishable pseudo-old feature space.

\vspace{2mm}
\noindent\textbf{Compatible learning with backfilling.}
Several studies~\cite{FCT, hot-refresh, RM, privacy,darwinian} have investigated compatible learning methods that enable online or low-cost backfilling for better performance. Several of these methods evaluate the quality of old embeddings in the gallery and assign backfilling priorities accordingly, aiming to improve the performance during the online backfilling process~\cite{hot-refresh,fastfill}. For example, RACT~\cite{hot-refresh} determines the backfilling priorities based on the uncertainty of the gallery embeddings. Besides, several methods~\cite{FCT,privacy,darwinian} opt to introduce a lightweight transformation module to map the old embeddings to the new feature space, \ie~low-cost backfilling. Some of these methods also choose to backfill only the embeddings with the poorest quality, aiming to further minimize computational costs~\cite{privacy, darwinian}.

While the above-mentioned methods effectively reduce the computational overhead of backfilling through feature transformations, chaining multiple learned mappings for successive model upgrades gradually increases computational costs over time and may eventually rival the expense of adopting a completely new model. Such a problem has also been discussed in BiCT~\cite{privacy}. In this paper, we focus on backfilling-free compatible learning and enhance the discriminative ability of the new model by adaptively relaxing the backward compatibility constraint through prototype perturbation during training.

\vspace{2mm}
\noindent\textbf{Lifelong learning.}
Lifelong learning~\cite{continual_learning,Zhou_2024_CVPR}, also known as incremental learning, aims to leverage the sequentially arriving data to update the model. Some early lifelong learning algorithms~\cite{continual_learning,chaudhry2018continual,de2021continual} do not consider model interaction over time; that is, the updated model is not required to maintain compatibility with previous versions. Recently, several studies~\cite{CL2R,biondi2024stationary,wan2022continual} have investigated the backward-compatible lifelong learning algorithms. Specifically, CL$^2$R~\cite{CL2R} and IAM-CL$^2$R~\cite{biondi2024stationary} utilize stationary representations learned by the d-Simplex fixed classifier to achieve backward compatibility of features across models updated in different incremental learning sessions. Besides class-incremental learning, some works have also explored lifelong person Re-ID~\cite{oh2024lifelong,cui2024learning} or lifelong natural image retrieval~\cite{wan2022continual} with backward compatibility. In addition, techniques similar to prototype perturbation, such as prototype augmentation, have been explored in lifelong learning to provide replay samples. The augmented prototypes in these methods~\cite{zhu2021prototype,kim2024cross} are employed to preserve the decision boundaries of previous tasks, whereas our prototype perturbation is designed to mitigate the influence of outdated and inaccurate discriminative boundaries.

\section{Methodology}

\begin{figure*}[t]
\centering
\includegraphics[width=1.0\textwidth]{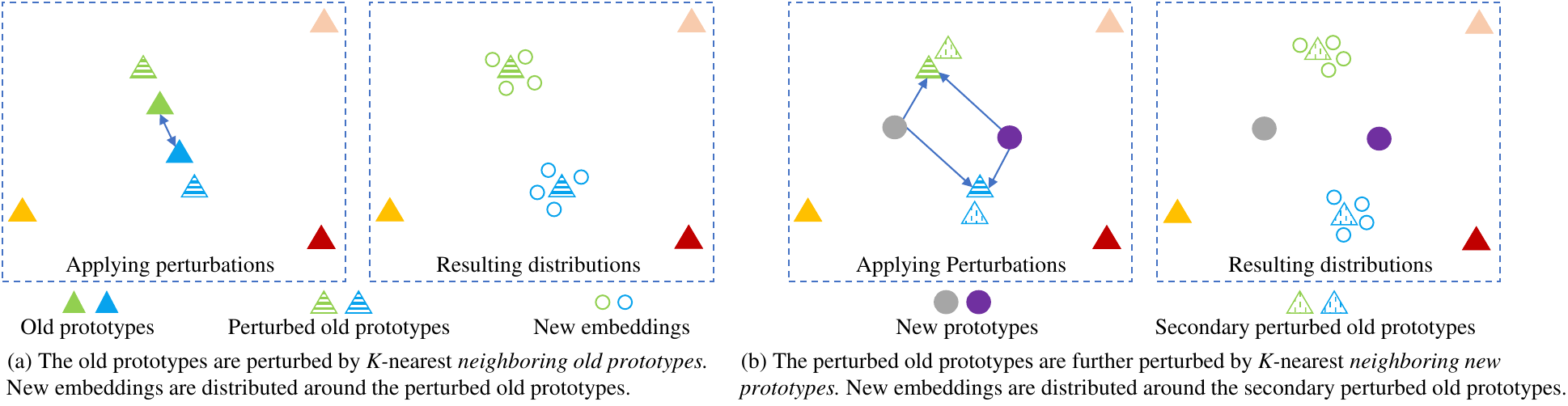}
\vspace{-6mm}
\caption{Illustration of neighbor-driven prototype perturbation. (a) Perturbations are calculated based on the neighboring old prototypes, and new embeddings are constrained to align with the perturbed old prototypes. (b) Perturbations are updated based on the neighboring new prototypes, and new embeddings are constrained to align with the secondary perturbed old prototypes.}
\label{Fig:illustration_NDPP}
\vspace{-4mm}
\end{figure*}

\subsection{Preliminaries}
\noindent\textbf{Problem formulation.}
Backward-Compatible Learning (BCL) aims to acquire a new embedding model compatible with the frozen old one. Denote the old and new models as $\phi_{o}\!:\!x\!\!\xrightarrow{}\!\! \mathbb{R}^{D_{o}}$ and $\phi_{n}\!:\!x\!\!\xrightarrow{}\!\! \mathbb{R}^{D_{n}}$, where $D_{o}$ and $D_{n}$ represent the dimensions of old and new embeddings, and $\phi_o$ and $\phi_n$ are trained on the datasets $\mathcal T_{o}$ and $\mathcal T_{n}$, respectively. Typically, $\mathcal T_{o}$ can be a subset of $\mathcal T_{n}$. Suppose we have a retrieval performance evaluation metric $\mathcal{M}(\phi_q, \phi_g, \mathcal{Q}, \mathcal{G})$, where $\mathcal Q$ and $\mathcal G$ are the query and gallery sets, and $\phi_q$ and $\phi_g$ are the models used to extract the embeddings of $\mathcal{Q}$ and $\mathcal{G}$, respectively. The new model $\phi_{n}$ is recognized to be backward-compatible with $\phi_{o}$ if it satisfies the empirical criterion $\mathcal{M}(\phi_{n}, \phi_{o}, \mathcal{Q}, \mathcal{G}) \!>\! \mathcal{M}(\phi_{o}, \phi_{o}, \mathcal{Q}, \mathcal{G})$~\cite{BCT}. From the above empirical criterion, it can be deduced that backward compatibility can be achieved as long as the new-to-old distance metric produces more accurate similarity rankings than the old-to-old one. The ideal case is that, given a new query embedding $\phi_n(x_i)$, the distance to the same-class gallery embedding $\phi_o(x_j)$ is less than the distance to any different-class gallery embedding $\phi_o(x_k)$. To this end, a robust new-to-old distance metric satisfying the following inequality is necessary~\cite{NCCL, DJAA}:
\begin{equation}
\label{Eq:BC_condition}
\begin{split}
    &d(\phi_n(x_i), \phi_o(x_j)) < d(\phi_n(x_i), \phi_o(x_k)), \\
    &{\forall}(i,j,k)\in\{(i,j,k):y_i=y_j\neq y_k\}.
\end{split}
\end{equation}
Herein $y_i$, $y_j$, and $y_k$ are the labels for samples $x_i$, $x_j$, and $x_k$, respectively. $d(\cdot, \cdot)$ denotes the distance metric. Please note that the above definition differs from that proposed in~\cite{BCT}. The definition in~\cite{BCT} requires the negative new-to-old pairs to remain at least as far apart as in the old model and the positive new-to-old pairs to remain at least as close as in the old model. It is a sufficient but not necessary condition for the empirical criterion. Moreover, when the number of classes increases, it becomes increasingly challenging to satisfy the requirement in~\cite{BCT} because the angles between class centers shrink, which has been empirically proven in~\cite{biondi2024stationary}.

Besides the backward compatibility, the discriminative ability of the new model is also crucial for BCL, which is usually evaluated by comparing its retrieval performance with that of a new model trained independently.

\vspace{2mm}
\noindent\textbf{Prototype-based contrastive learning for BCL.}
Contrastive learning is typically introduced to learn a robust new-to-old distance metric in BCL, including the point-to-point (P2P)~\cite{hot-refresh,Online-Backfilling} and point-to-set (P2S)~\cite{DualTuning} contrastive learning methods. The P2P method tends to be sensitive to outliers in the old feature space. In contrast, the P2S method considers a set of samples from the same class as positive or negative examples. The influence of the individual outlier sample is diluted by the sample set, and thus P2S contrastive learning deals with the outliers more robustly. Hence, we prefer P2S contrastive learning as our baseline method.

We perform P2S contrastive learning with the prototype of each class, which is the average of the embeddings belonging to that class, \ie~the class center. During BCL, we use the frozen old model to extract the embeddings of samples in $\mathcal T_{n}$ and then compute the prototypes of all classes. They together represent the feature distribution of the old model and are referred to as \emph{old prototypes}. Denoting the old prototype of class $c$ by $p_{o}^{c}$, the prototype-based contrastive loss for training the new model $\phi_n$ can be formulated as:
\begin{equation}
\label{Eq: prototype_contrastive}
\setlength{\abovedisplayskip}{6pt}
\setlength{\belowdisplayskip}{6pt}
\mathcal{L}_{bc} = -\log \frac{{\exp(\langle e_n^c , p^c_o\rangle / \tau)}}{{\sum_{{c'\in \mathcal{C}}} \exp(\langle e_n^c, p^{c'}_o \rangle / \tau)}},
\end{equation}
where $\langle \cdot, \cdot \rangle$ denotes cosine similarity, $e_n^c=\phi_n(x^c)$, $x^c\! \in\! \mathcal T_{n}$ is the sample belonging to class $c$, $\mathcal{C}$ is the class set of $\mathcal T_{n}$, and $\tau$ is the temperature factor.

Following existing approaches~\cite{BCT,advbct}, we formulate the discriminative learning of the new model as a classification task. Specifically, we build a fully connected layer on the embedding model and introduce a cross-entropy loss $\mathcal{L}_{ce}$ for classification. Hence, the overall training objective of the new model can be formulated as:
\begin{equation}
\label{eq:total_loss}
\setlength{\abovedisplayskip}{6pt}
\setlength{\belowdisplayskip}{6pt}
\mathcal{L} = \mathcal{L}_{ce} + \lambda\mathcal{L}_{bc}.
\end{equation}
Herein $\lambda$ is a balance weight for $\mathcal{L}_{bc}$.

\subsection{Overview of prototype perturbation}
Although prototype-based contrastive learning addresses the outlier issue, it still suffers from the adverse effect of the indistinguishable classes in the old feature space, due to the strict alignment constraint formulated by Eq.~\eqref{Eq: prototype_contrastive}. To alleviate this adverse effect, we propose the prototype perturbation mechanism to generate a pseudo-old feature distribution as the backward-compatible learning target. Although the new embeddings are constrained to surround the pseudo-old prototypes, Eq.~\eqref{Eq:BC_condition} can still be satisfied as long as the perturbation is appropriately determined. Technically, the key to prototype perturbation is how to properly push the old prototypes away from their indistinguishable neighbors.

In light of this idea, we devise two approaches to determine the perturbations: Neighbor-Driven Prototype Perturbation (NDPP) and Optimization-Driven Prototype Perturbation (ODPP). NDPP and ODPP will produce the pseudo-old prototypes for calculating the prototype-based contrastive loss. Namely, our two approaches will update Eq.~\eqref{Eq: prototype_contrastive} to improve the backward-compatible learning performance. In the following, we detail the NDPP and ODPP approaches and present how to leverage the pseudo-old prototypes to perform backward-compatible learning.

\vspace{-2mm}
\subsection{Neighbor-driven prototype perturbation}
NDPP calculates the perturbation for an old prototype based on its similarity with the neighbors. The basic principle is that each old prototype is presumed to experience repulsion from its neighbors, and the intensity of the repulsion is proportional to their similarity. Specifically, we consider the neighbors from both the old and new feature spaces for each old prototype to be perturbed. Next, we present the prototype perturbation based on different neighbors.

\vspace{2mm}
\noindent\textbf{Prototype perturbation based on old neighbors.} Figure~\ref{Fig:illustration_NDPP} (a) illustrates the prototype perturbation scheme based on old neighboring prototypes. For the old prototype $p^{c}_{o}$ of class $c$, we first find its $K$-nearest neighbors by calculating its cosine similarities with the old prototypes of the other classes. We denote the $K$-nearest neighbors by $\{p_o^{c'}\}_{c' \in \mathcal{K}_{o}^{c}}$ and the corresponding similarities by $\{s_{o2o}^{c'}\}_{c' \in \mathcal{K}_{o}^{c}}$, where $\mathcal{K}_{o}^{c}$ denotes the class set of these neighbors. We calculate the perturbation $r^{c}_{o}$ for the old prototype $p_{o}^{c}$ by:
\begin{equation}
\label{Eq:perturbation_old}
\setlength{\abovedisplayskip}{6pt}
\setlength{\belowdisplayskip}{6pt}
r^{c}_{o} = \frac{\sum_{c' \in \mathcal{K}_{o}^{c}}s_{o2o}^{c'}(p^{c}_{o}-p^{c'}_{o})}{\sum_{c' \in \mathcal{K}_{o}^{c}}(s_{o2o}^{c'})}.
\end{equation}
Herein $r^{c}_{o}$ is calculated by summarizing the repulsion between $p_{o}^{c}$ and its neighbors using the corresponding similarity as the weight, and the repulsion is defined by the vector $p^{c}_{o}-p^{c'}_{o}$, which points to $p^{c'}_{o}$ from $p^{c}_{o}$. Applying such a perturbation to the real old prototype $p_{o}^{c}$, the resulting pseudo-old prototype $\hat p_{o}^{c}$ can be formulated by:
\begin{equation}
\label{Eq:add_perturbation}
\setlength{\abovedisplayskip}{6pt}
\setlength{\belowdisplayskip}{6pt}
\hat p_{o}^{c} = p_{o}^{c} + \alpha_{1} r^{c}_{o},
\end{equation}
where $\alpha_{1}$ is a factor rescaling the perturbation amplitude.

Afterward, we use the pseudo-old prototypes to replace the real old ones in the contrastive loss for backward-compatible learning. In particular, we use the pseudo-old prototypes in the positive pairs but reserve the real old prototypes in the negative pairs. Thus, the prototype-based contrastive loss, Eq.~\eqref{Eq: prototype_contrastive}, is updated as:
\begin{equation}
\label{Eq:perturbation_o2o}
\setlength{\abovedisplayskip}{6pt}
\setlength{\belowdisplayskip}{6pt}
\mathcal{L}_{bc} \!=\! -\!\log\! \frac{{\exp(\langle e_n^c, \hat p^c_o\rangle / \tau)}}{{\exp(\langle e_n^c, \hat p^c_o \rangle / \tau)}\!+\!\sum\limits_{{c'\in \mathcal{C}\backslash c}}\! \exp(\langle e_n^c, p^{c'}_{o} \rangle / \tau)}.
\end{equation}

\vspace{2mm}
\noindent\textbf{Prototype perturbation based on joint neighbors.}
Prototype perturbation aims to relax the alignment constraint properly to facilitate the discriminative learning of the new model. Nevertheless, the above perturbations based on only the old neighbors overlook the distribution of the new features, being independent of the training of the new model. To overcome this limitation, we propose to leverage the \emph{new prototypes} calculated by the new model to further update the pseudo-old prototypes. As shown in Figure~\ref{Fig:illustration_NDPP} (b), such a design aims to acquire the secondary perturbed old prototypes that are more effective in enhancing the discriminative ability of the new model during BCL.

Over the training process, we update the pseudo-old prototypes continuously along with the evolution of the new embedding model. Specifically, at the beginning of each training epoch, we first obtain the new prototypes $\phi_{n}^{c}$ of each class by computing the class center of the new embeddings, which will be used to repel the nearby pseudo-old prototypes. For the pseudo-old prototype $\hat p^{c}_{o}$, we find its $K$-nearest neighboring new prototypes, denoted by $\{p^{c'}_{n}\}_{c' \in \mathcal{K}_{n}^{c}}$, and the corresponding similarities, denoted by $\{s_{n2o}^{c'}\}_{c' \in \mathcal{K}_{n}^{c}}$, where $\mathcal{K}_{n}^{c}$ denotes the class set of these neighboring new prototypes. The new-to-old perturbation $r^{c}_{n}$ for the pseudo-old prototype $\hat p^{c}_{o}$ can be formulated as:
\begin{equation}
\label{Eq:perturbation_new}
r^{c}_{n} = \frac{\sum_{c' \in \mathcal{K}_{n}^{c}}s_{n2o}^{c'}(\hat p^{c}_{o}-p^{c'}_{n})}{\sum_{c' \in \mathcal{K}_{n}^{c}}(s_{n2o}^{c'})}.
\end{equation}
We denote $r^{c}_{n}$ calculated at the $t^{th}$ epoch by $r^{c}_{t}$. Thus the pseudo-old prototype $\hat p^{c}_{o\_t}$ with a secondary perturbation at the $t^{th}$ epoch is:
\begin{equation}
\label{Eq:add_n2o_perturbation}
\setlength{\abovedisplayskip}{6pt}
\setlength{\belowdisplayskip}{6pt}
\hat p_{o\_t}^{c} = p_{o}^{c} + \alpha_1 r^{c}_{o} +\alpha_2 r^{c}_t,
\end{equation}
where $\alpha_2$ is a factor rescaling the amplitude of $r^{c}_t$. We replace $\hat p_{o}^{c}$ with $\hat p_{o\_t}^{c}$ to calculate the contrastive loss formulated by Eq.~\eqref{Eq:perturbation_o2o} at the $t^{th}$ epoch.
By continuously updating the pseudo-old prototypes conditioned on the changing feature distribution of the new model, we can adaptively relax the alignment constraints during training, further improving the discriminative ability of the new model. Algorithm~\ref{algorithm1} summarizes the training process of NDPP.

\begin{algorithm}[t!]
\caption{Training process of NDPP}
    \label{algorithm1}
    \SetAlgoNlRelativeSize{0}
    \SetKwInOut{Require}{Require}
    \SetAlgoLined
    \LinesNumbered
    \SetNoFillComment
    \SetCommentSty{}
    \SetSideCommentRight
    \Require{Training set $\mathcal{D}$, models $\phi_o$ and $\phi_n$;}
    \tcp{\small \color{gray} Prepare for training}
    {$\{p_o^c\}_{c\in\mathcal{C}} \gets$ GetPrototypes($\phi_o$, $\mathcal{D}$)\;}
    \For{$p_o^c \in \{p_o^c\}_{c\in\mathcal{C}}$}{
        {$K_{o2o}$, $S_{o2o}\gets$ FindNeighbors($p_o^c,\{p_o^c\}_{c\in\mathcal{C}}$)\;}
        {$r^{c}_{o}\gets\frac{\sum_{c' \in \mathcal{K}_{o}^{c}}s_{o2o}^{c'}(p^{c}_{o}-p^{c'}_{o})}{\sum_{c' \in \mathcal{K}_{o}^{c}}(s_{o2o}^{c'})}$\;}
        {$\hat{p_o^c} \gets p_o^c + \alpha_1 \* r_o^c$\;}
    }
    {$P'_o\gets \{\hat{p_o^c}\}$\;}
    \tcp{\small \color{gray} Training loops}
    \For{\textnormal{each epoch}}{
        {$\{p_n^c\}_{c\in\mathcal{C}} \gets$ GetPrototypes($\phi_n$)\;}
        \For{$\hat{p_o^c} \in P'_o$}{
        {$K_{n2o}$, $S_{n2o}\gets$ FindNeighbors($p_o^c$, $\{p_n^c\}_{c\in\mathcal{C}}$)\;}
        {$r^{c}_{n}\gets\frac{\sum_{c' \in \mathcal{K}_{n}^{c}}s_{n2o}^{c'}(\hat p^{c}_{o}-p^{c'}_{n})}{\sum_{c' \in \mathcal{K}_{n}^{c}}(s_{n2o}^{c'})}$\;}
        {$\hat{p_o^c} \gets \hat{p_o^c} + \alpha_2 \* r_n^c$\;}
        }
        \For{\textnormal{each iteration}}{
            {$\mathcal{B}\gets$ SampleData($\mathcal{D}$)\;}
            {$g\gets$ $\frac{\partial\mathcal{L}(\mathcal{B},\phi_n,P'_o)}{\partial\phi_n}$\;}
            {Update $\phi_n$ by $g$\;}
        }
    }
\end{algorithm}

\vspace{-2mm}
\subsection{Optimization-driven prototype perturbation}
Besides NDPP which calculates the perturbation heuristically, we also designed another implementation for our prototype perturbation mechanism, ODPP. ODPP introduces a learnable perturbation vector $r^{c}_l$ for each old prototype $p_{o}^{c}$, and then minimizes the similarities between the indistinguishable prototypes~\wrt~the learnable perturbations. Similar to NDPP, we take the feature distributions of both the old and new models into account for designing the objective function. In the following, we present how to learn the perturbations based on different prototypes.

\vspace{2mm}
\noindent\textbf{Learning perturbation based on old prototypes.}
Given that only the prototype pairs being hard to distinguish need to be perturbed, we employ a hinge loss as the objective:
\begin{equation}
\label{Eq:odpp_old}
\setlength{\abovedisplayskip}{6pt}
\setlength{\belowdisplayskip}{6pt}
\mathcal{L}_{ptb\_o} = \sum_{c\in\mathcal{C}}\sum_{c'\in\mathcal{C}\backslash c} {\rm max}(0,(p_{o}^{c}+r_{l}^{c})\cdot(p_{o}^{c'}+r_{l}^{c'})-\theta_{old}),
\end{equation}
where $\theta_{old}$ is the threshold identifying the old prototype pair being hard to distinguish, and $\cdot$ denotes the inner product. The class set $\mathcal C$ typically contains thousands of classes at least, and thus we cannot enumerate all pairs of $(c, c')$ to optimize Eq.~\eqref{Eq:odpp_old}. Technically, we adopt mini-batch Stochastic Gradient Descent (SGD) to minimize $\mathcal{L}_{ptb\_o}$.
After the optimization, the pseudo-old prototype $\hat p_{o}^{c}$ is calculated by:
\begin{equation}
\label{Eq:add_odpp_perturbation}
\hat p_{o}^{c} = p_{o}^{c} + r_{l}^{c}.
\end{equation}
Note that no scale factor is involved in Eq.~\eqref{Eq:add_odpp_perturbation}. Then we could calculate the contrastive loss formulated by Eq.~\eqref{Eq:perturbation_o2o} with the above pseudo-old prototype.

\vspace{2mm}
\noindent\textbf{Learning perturbation based on joint prototypes.}
Similar to NDPP, we can also include the new prototypes in the optimization of the learnable perturbations during BCL. Thus the objective function can be modified as:
\begin{equation}
\begin{split}
\label{Eq:odpp_new}
\mathcal{L}_{ptb} = & \sum_{c\in\mathcal{C}}\sum_{c'\in\mathcal{C}\backslash c} {\rm max}(0,(p_{o}^{c}\!+\!r_{l}^{c})\!\cdot\!(p_{o}^{c'}\!+\!r_{l}^{c'})\!-\!\theta_{old}) \\ &+\gamma\sum_{c\in\mathcal{C}}\sum_{c''\in\mathcal{C}\backslash c}{\rm max}(0,(p_{o}^{c}\!+\!r_{l}^{c})\!\cdot\! p_{n}^{c''}\!-\!\theta_{new}),
\end{split}
\end{equation}
where $\gamma$ is a balance weight for the second term, and $\theta _{new}$ is the threshold identifying the similar old-to-new prototype pair.
We optimize Eq.~\eqref{Eq:odpp_new} at the beginning of each training epoch during BCL and use the resulting perturbations to generate the pseudo-old prototypes for BCL. Algorithm~\ref{algorithm_odpp} describes the training process of ODPP.

\begin{algorithm}[t!]
    \caption{Training process of ODPP}
    \label{algorithm_odpp}
    \SetAlgoNlRelativeSize{0}
    \SetKwInOut{Require}{Require}
    \SetAlgoLined
    \LinesNumbered
    \SetNoFillComment
    \SetCommentSty{}
    \SetSideCommentRight
    \Require{Training set $\mathcal{D}$, models $\phi_o$ and $\phi_n$;}
    \tcp{\small \color{gray} Prepare for training}
    {$\{p_o^c\}_{c\in\mathcal{C}} \gets$ GetPrototypes($\phi_o$, $\mathcal{D}$)\;}
    {$r_l^c\gets \underset{r_l^c}{\arg\min}~\mathcal{L}_{ptb\_o}$ \tcp*{\small \color{gray} Get perturbations}}
    {$\{\hat{p_o^c}\}_{c\in\mathcal{C}}\gets \{p_o^c\}_{c\in\mathcal{C}} + r_l^c$\;}
    \tcp{\small \color{gray} Training loops}
    \For{\textnormal{each epoch}}{
        {$\{p_n^c\}_{c\in\mathcal{C}} \gets$ GetPrototypes($\phi_n$, $\mathcal{D}$)\;}
        {$r_l^c\gets \underset{r_l^c}{\arg\min}~\mathcal{L}_{ptb}$ \tcp*{\small \color{gray} Update perturbations}}
        {$\{\hat{p_o^c}\}_{c\in\mathcal{C}}\gets \{p_o^c\}_{c\in\mathcal{C}} + r_l^c$\;}
        \For{\textnormal{each iteration}}{
            {$\mathcal{B}\gets$ SampleData($\mathcal{D}$)\;}
            {$g\gets$ $\frac{\partial\mathcal{L}(\mathcal{B},\phi_n,P'_o)}{\partial\phi_n}$\;}
            {Update $\phi_n$ by $g$\;}
        }
    }
\end{algorithm}

Similar to NDPP, ODPP obtains perturbations also based on the similarity between prototypes. NDPP and ODPP differ primarily in the scope of the feature space used for computing prototype perturbations, as well as in computational complexity. Specifically, NDPP directly calculates the perturbation for each prototype by aggregating the repulsion from its $K$-nearest neighbors. Namely, the perturbation is derived from a local region of the feature space. Hence, NDPP offers relatively low computational complexity in obtaining perturbations; however, the resulting perturbations tend to be suboptimal, as they are primarily determined by local information. By contrast, ODPP minimizes Eq. (11) by mini-batch SGD to learn appropriate perturbations pushing similar prototypes apart. That is, ODPP refines the perturbations based on the overall feature space iteratively, finally yielding a solution close to the global optimal perturbations. However, ODPP involves greater computational complexity than NDPP due to the necessity of multiple backpropagation iterations.

In practical applications, NDPP is recommended for prioritization under two specific conditions: (1) when it is uncomplicated to obtain suitable perturbations, such as in cases where the training dataset contains a limited number of classes (resulting in higher inherent separability between classes); and (2) when the training efficiency of the new model is a priority. On the other hand, ODPP is preferable when the training efficiency of the new model is not a primary concern and obtaining suitable perturbations is more challenging, for example, in cases where the training dataset contains a large number of classes.

\section{Experiments}
\begin{table}[t!]
\centering
\setlength{\tabcolsep}{2pt}
\renewcommand{\arraystretch}{1.2}
\caption{Data allocations for different BCL settings. R18 and R50 denote ResNet18 and ResNet50, respectively.}
\vspace{-2mm}
\resizebox{1.0\linewidth}{!}{
\begin{tabular}{llccccc}
\toprule
 \multirow{2}{*}{Dataset} &\multirow{2}{*}{\shortstack{Setting}} & \multicolumn{2}{c}{Old Training Set} && \multicolumn{2}{c}{New Training Set} \\ 
\cline{3-4} \cline{6-7}  
                     && images         & classes          && images         & classes           \\ 
\midrule
\multirow{3}{*}{GLDv2} & Data-Extension (9\%$\rightarrow$30\%)  &142,772    &7,318   && 470,369    & 24,393  \\
& Data-Extension (30\%$\rightarrow$100\%) & 470,369 & 24,393 && 1,580,470 & 81,313 \\
& Backbone-extension (R18$\rightarrow$R50)  &142,772    &7,318   &&470,369    &24,393   \\
\midrule
\multirow{2}{*}{In-shop} & Data-Extension (30\%$\rightarrow$100\%)  & 7,525    &1,199   &&25,880   &3,997  \\
& Backbone-extension (R18$\rightarrow$R50)  & 7,525    &1,199   &&25,880    &3,997   \\
\midrule
\multirow{2}{*}{Market-1501} & Data-Extension (50\%$\rightarrow$100\%) & 5,663 & 312 && 12,936 & 751 \\
& Data-Extension (10\%$\rightarrow$100\%) & 1,371 & 75 && 12,936 & 751 \\
\midrule
RSTPReid & Data-Extension (50\%$\rightarrow$100\%) & 9,255 & 1,851&& 18,505& 3,701\\
\bottomrule
\end{tabular}}
\label{Tab:datasets}
\vspace{-3mm}
\end{table}

\subsection{Experiment settings}
Collecting more training data and expanding the backbone scale are typical methods for model upgrade. Hence, we experiment with two settings: 1) \emph{Data-extension}, where more data are used to train a new model with the same backbone as the old one; 2) \emph{Backbone-extension}, where more data are used to train a new model with a larger backbone than the old one. Herein, we opt for extending the training data by introducing new classes, which can enhance the diversity of classes and is more aligned with real-world applications. Next, we detail the datasets and metrics.

\begin{table*}[t!]
\centering
\setlength{\tabcolsep}{9pt}
\renewcommand{\arraystretch}{1.1}
\caption{Results on RParis, ROxford, and GLDv2-test in the single-step BCL experiment. $\dagger$ denotes that the model is trained independently. $\mathcal{P}_{up}$, $\mathcal{P}_{comp}$, and $\mathcal{P}_{1-score}$ are calculated based on the mAP over RParis, ROxford, and GLDv2-test.}
\vspace{-2mm}
\resizebox{0.95\linewidth}{!}{
\begin{tabular}{llcccccccccccc}
\toprule
\multirow{2}{*}{\shortstack{Allocation\\Type}} &\multirow{2}{*}{Methods} & \multicolumn{2}{c}{RParis (mAP)} && \multicolumn{2}{c}{ROxford (mAP)} && \multicolumn{2}{c}{GLDv2-test (mAP)} &\multirow{2}{*}{$\mathcal{P}_{up}$} &\multirow{2}{*}{$\mathcal{P}_{comp}$} &\multirow{2}{*}{$\mathcal{P}_{1-score}$}\\ 
\cline{3-4} \cline{6-7} \cline{9-10} 
                     & & self-test         & cross-test          && self-test         & cross-test  && self-test         & cross-test         \\ 
\midrule
            \multirow{12}{*}{\shortstack{Data-\\Extension\\(9\%$\rightarrow$30\%)}} 
            & \multicolumn{1}{>{\cellcolor[gray]{0.9}}l}{$\dagger$R18 (old)}  
            & \multicolumn{1}{>{\cellcolor[gray]{0.9}}c}{67.31}  
            & \multicolumn{1}{>{\cellcolor[gray]{0.9}}c}{--}  
            & \multicolumn{1}{>{\cellcolor[gray]{0.9}}c}{}  
            & \multicolumn{1}{>{\cellcolor[gray]{0.9}}c}{41.82}  
            & \multicolumn{1}{>{\cellcolor[gray]{0.9}}c}{--}  
            & \multicolumn{1}{>{\cellcolor[gray]{0.9}}c}{}  
            & \multicolumn{1}{>{\cellcolor[gray]{0.9}}c}{7.30}  
            & \multicolumn{1}{>{\cellcolor[gray]{0.9}}c}{--}  
            & \multicolumn{1}{>{\cellcolor[gray]{0.9}}c}{--}  
            & \multicolumn{1}{>{\cellcolor[gray]{0.9}}c}{--}  
            & \multicolumn{1}{>{\cellcolor[gray]{0.9}}c}{--} \\
            & \multicolumn{1}{>{\cellcolor[gray]{0.9}}l}{$\dagger$R18 (new)}  
            & \multicolumn{1}{>{\cellcolor[gray]{0.9}}c}{75.08}  
            & \multicolumn{1}{>{\cellcolor[gray]{0.9}}c}{--}  
            & \multicolumn{1}{>{\cellcolor[gray]{0.9}}c}{}  
            & \multicolumn{1}{>{\cellcolor[gray]{0.9}}c}{55.77}  
            & \multicolumn{1}{>{\cellcolor[gray]{0.9}}c}{--}  
            & \multicolumn{1}{>{\cellcolor[gray]{0.9}}c}{}  
            & \multicolumn{1}{>{\cellcolor[gray]{0.9}}c}{12.08}  
            & \multicolumn{1}{>{\cellcolor[gray]{0.9}}c}{--}  
            & \multicolumn{1}{>{\cellcolor[gray]{0.9}}c}{--}  
            & \multicolumn{1}{>{\cellcolor[gray]{0.9}}c}{--}  
            & \multicolumn{1}{>{\cellcolor[gray]{0.9}}c}{--} \\
            & BCT~\cite{BCT}  &71.52       & 68.37       && 51.16       & 44.75  &&10.38  & 8.72 &47.75  &55.34    &51.23   \\
            & UniBCT~\cite{unibct}  & 73.94       & 69.52       && 50.04       & 43.79 &&\underline{12.60}  &8.72  &49.38   &55.99 &52.47 \\ 
            & RACT~\cite{hot-refresh}  & 74.91      & 67.87      &&54.35       & 41.55 &&11.45&  6.63 &49.33&49.27&49.29 \\         
            & AdvBCT~\cite{advbct}  & 74.24       & 68.70       && 54.08      & 42.88 &&11.57  & 7.94 &49.30&53.23&51.19 \\ 
            & Dual-Tuning~\cite{DualTuning} &73.43 & 69.40&&53.97&45.05&&11.19&\underline{8.75}&48.93&56.66&52.51 \\
            &SSPL-CE~\cite{SSPL}&72.67&68.97&&50.90&45.09&&11.04&8.12&48.29&55.14&51.49\\
            &SSPL~\cite{SSPL} & 67.42 &67.05 &&42.77&42.07&&8.14&7.58&44.52&50.36&47.20\\
            & BT$^2$~\cite{bt2}& 73.20 & 68.94 && 52.93 &45.96&&11.34&8.34&48.86&56.00&52.18\\
            & \textbf{NDPP} (Ours)  & \underline{76.21}      &  \textbf{71.10}      && \textbf{58.02}       & \textbf{46.42} &&\textbf{12.67}  &\textbf{8.88}  & \textbf{50.87} & \textbf{59.44} & \textbf{54.80} \\     
            & \textbf{ODPP} (Ours)  & \textbf{76.84}       & \underline{70.99}      && \underline{56.45}       & \underline{46.24} && 12.44 & 8.60 & \underline{50.55} & \underline{58.75} & \underline{54.32} \\
\midrule
            \multirow{12}{*}{\shortstack{Backbone-\\Extension\\(R18$\rightarrow$R50)}} 
            & \multicolumn{1}{>{\cellcolor[gray]{0.9}}l}{$\dagger$R18 (old)}  
            & \multicolumn{1}{>{\cellcolor[gray]{0.9}}c}{67.31}  
            & \multicolumn{1}{>{\cellcolor[gray]{0.9}}c}{--}  
            & \multicolumn{1}{>{\cellcolor[gray]{0.9}}c}{}  
            & \multicolumn{1}{>{\cellcolor[gray]{0.9}}c}{41.82}  
            & \multicolumn{1}{>{\cellcolor[gray]{0.9}}c}{--}  
            & \multicolumn{1}{>{\cellcolor[gray]{0.9}}c}{}  
            & \multicolumn{1}{>{\cellcolor[gray]{0.9}}c}{7.30}  
            & \multicolumn{1}{>{\cellcolor[gray]{0.9}}c}{--}  
            & \multicolumn{1}{>{\cellcolor[gray]{0.9}}c}{--}  
            & \multicolumn{1}{>{\cellcolor[gray]{0.9}}c}{--}  
            & \multicolumn{1}{>{\cellcolor[gray]{0.9}}c}{--} \\
            & \multicolumn{1}{>{\cellcolor[gray]{0.9}}l}{$\dagger$R50 (new)}  
            & \multicolumn{1}{>{\cellcolor[gray]{0.9}}c}{81.41}  
            & \multicolumn{1}{>{\cellcolor[gray]{0.9}}c}{--}  
            & \multicolumn{1}{>{\cellcolor[gray]{0.9}}c}{}  
            & \multicolumn{1}{>{\cellcolor[gray]{0.9}}c}{63.59}  
            & \multicolumn{1}{>{\cellcolor[gray]{0.9}}c}{--}  
            & \multicolumn{1}{>{\cellcolor[gray]{0.9}}c}{}  
            & \multicolumn{1}{>{\cellcolor[gray]{0.9}}c}{15.40}  
            & \multicolumn{1}{>{\cellcolor[gray]{0.9}}c}{--}  
            & \multicolumn{1}{>{\cellcolor[gray]{0.9}}c}{--}  
            & \multicolumn{1}{>{\cellcolor[gray]{0.9}}c}{--}  
            & \multicolumn{1}{>{\cellcolor[gray]{0.9}}c}{--} \\
            & BCT~\cite{BCT}  &77.20       & 70.28       && 57.58       & 46.27  &&12.53  &9.53    &47.23&55.73&51.11  \\
            & UniBCT~\cite{unibct}  & 77.71       & 71.34       && 58.39       & 45.96 && 13.49 &9.38  &47.91&56.07&51.66 \\ 
            &RACT~\cite{hot-refresh}  & 78.30       &67.52       && 60.10       & 44.04 && 13.27 & 8.80 &48.07&52.51&50.16\\      
            & AdvBCT~\cite{advbct}  & \underline{79.50}       & 70.32       && 61.12       & 44.11  && 14.09 & 9.42&48.77&54.82&51.60\\ 
            & Dual-Tuning~\cite{DualTuning} & 76.93 & 70.85&&57.85&46.66&&13.42&9.04&47.72&55.71&51.41\\
            &SSPL-CE~\cite{SSPL}&76.26&70.58&&54.61&45.87&&12.70&9.12&46.84&55.33&50.73\\
            & SSPL~\cite{SSPL} &67.18&67.36&&40.71&41.59&&8.19&7.27&41.75&49.91&45.41 \\
            & BT$^2$~\cite{bt2} &75.50&69.84&&53.46&45.45&&12.42&8.37&46.46&53.98&49.93\\
            & \textbf{NDPP} (Ours)  & \textbf{79.55}       &  \underline{71.42}      && \textbf{62.98}       & \underline{49.25}  &&\underline{14.99}  &\underline{9.65} &\textbf{49.51}&\underline{57.63}&\underline{53.26}\\ 
            & \textbf{ODPP} (Ours)  & \textbf{79.55}       & \textbf{71.96}      && \underline{62.46}       & \textbf{49.40} && \textbf{15.10} & \textbf{10.06} &\underline{49.50}&\textbf{58.41}&\textbf{53.59} \\   
\bottomrule
\end{tabular}}
\label{Tab:gldv2}
\vspace{-3mm}
\end{table*}

\vspace{2mm}
\noindent\textbf{Datasets.} (1) \textbf{GLDv2}~\cite{gldv2} is comprised of large-scale images representing human-made and natural landmarks. The clean version of GLDv2 is composed of 1,580,470 images from 81,313 landmarks. Considering it contains substantial training data, we conduct both the regular \emph{single-step backward-compatible learning} experiment and the \emph{sequential backward-compatible learning} experiment on GLDv2. Particularly, we gradually increase the training classes from 9\% to 30\% and further to 100\% to train the models for the data-extension settings. Each new model is constrained to be compatible with its previous version. For the models trained on GLDv2, we evaluate their performance on RParis~\cite{rparis&roxford}, ROxford~\cite{rparis&roxford}, and GLDv2-test~\cite{gldv2}.
(2) \textbf{In-shop}~\cite{inshop} contains 52,712 in-shop clothes images from 7,982 clothing items. We split the dataset into a training set with 3,997 categories and a testing set with 3,985 categories. Considering its limited data size, we only conduct the single-step backward-compatible learning experiment on the In-shop dataset. To be specific, we randomly sample 30\% of the training classes to train the old model and use the entire training set to train a new model with backward compatibility. (3) \textbf{Market-1501}~\cite{Market1501} is a large-scale person Re-ID dataset. It contains 1,501 distinct identities, including 751 identities for training and 750 identities for testing. Following RM~\cite{RM}, we use 50\% and 100\% of data to train the old and new models, respectively. To further compare with RBCL~\cite{RBCL}, we follow the In-Domain Supervised Setting 1 (ID-S-1) in RBCL, using random 10\% classes to train the old model. Besides, we also conduct the sequential BCL experiment on Market-1501 following the setting of CoReS~\cite{Cores}. (4) \textbf{RSTPReid}~\cite{RSTPReid} is a text-based person Re-ID dataset containing 3701 identities for training. We use this dataset to assess the effectiveness of our method for multimodal retrieval. We use 50\% and 100\% of data to train the old and new models, respectively. Table~\ref{Tab:datasets} details the data allocations.

\newcommand{\tabincell}[2]{\begin{tabular}{@{}#1@{}}#2\end{tabular}}

\vspace{2mm}
\noindent\textbf{Metrics.} 
We employ cosine similarity to rank the retrieved images and then calculate mean Average Precision (mAP) as the metric. To evaluate the performance of BCL algorithms, we report the \emph{self-test} performance $\mathcal M(\phi_{o}, \phi_{o}, \mathcal{Q}, \mathcal{G})$ or $\mathcal M(\phi_{n}, \phi_{n}, \mathcal{Q}, \mathcal{G})$; and the \emph{cross-test} performance $\mathcal M(\phi_{n}, \phi_{o}, \mathcal{Q}, \mathcal{G})$. Following AdvBCT~\cite{advbct}, we also use $\mathcal{P}_{comp}$, $\mathcal{P}_{up}$ and $\mathcal{P}_{1-score}$ calculated based on mAP as metrics. Specifically, $\mathcal{P}_{comp}$ and $\mathcal{P}_{up}$ measure the compatibility and discriminative ability of the new model, respectively. $\mathcal{P}_{1-score}$ takes both $\mathcal{P}_{comp}$ and $\mathcal{P}_{up}$ into account, which is used to measure the overall performance. We also use the \textit{Compatibility Matrix}, the \textit{Average Multi-model Compatibility} (\textit{AC}), the \textit{Average Multi-model Accuracy} (\textit{AM})~\cite{Cores} to assess sequential model upgrade performance.

\subsection{Implementation details}
For the visual retrieval task, we opt for ResNet~\cite{ResNet} pretrained on ImageNet~\cite{Imagenet} as the backbone of the embedding model, unless otherwise specified. We employ an MLP to project the embeddings to the feature space with a dimension of 256, similar to AdvBCT~\cite{advbct}. For the multimodal retrieval task, we use the pretrained APTM~\cite{APTM} as the embedding model, which also projects the input text or image into an embedding space of 256 dimensions. $\tau$ is set to 0.07. $\lambda$ is set to 1.0. On GLDv2, we train the models for 30 epochs using SGD with a learning rate beginning from 0.1 and dropping by 10 at the $5^{th}$, $10^{th}$, and $20^{th}$ epochs. On In-shop, we train the models for 200 epochs using SGD with a learning rate beginning from 0.1 and dropping by 10 at the $30^{th}$, $60^{th}$, and $120^{th}$ epochs. The weight decay is set to 5e-4, and the momentum is set to 0.9 for both GLDv2 and In-shop. On Market-1501, we train the models for 120 epochs using Adam~\cite{Adam} with a learning rate beginning from $3.5\times10^-5$ and dropping by 10 at $40^{th}$ and $90^{th}$ epochs. On RSTReid, we follow the finetuning configuration of APTM~\cite{APTM} to train the new model. For ODPP, we minimize Eq.~\ref{Eq:odpp_new} to optimize the learnable perturbation at the beginning of each training epoch of the new model. The optimization is configured with 50 epochs, a learning rate of 0.001, and a batch size of 1024. For each epoch, we randomly sample $N_{img}$ pairs of prototypes for training, where $N_{img}$ is the training dataset scale.

\subsection{Comparison with state-of-the-art algorithms}
We compare our NDPP and ODPP with state-of-the-art algorithms on diverse retrieval datasets. The methods involved in the comparison include BCT~\cite{BCT}, UniBCT~\cite{unibct}, RACT~\cite{hot-refresh}, AdvBCT~\cite{advbct}, CoReS~\cite{Cores}, Dual-Tuning~\cite{DualTuning}, SSPL~\cite{SSPL}, BT$^2$~\cite{bt2}, RM~\cite{RM}, and RBCL~\cite{RBCL}. In our experiment, we use the implementation of BCT, UniBCT, RACT, and AdvBCT released in the official code\footnote{https://github.com/Ashespt/AdvBCT} of~\cite{advbct}. For Dual-Tuning, SSPL, and BT$^2$, we use the official codes\footnote{https://github.com/YifeiZhou02/BT-2}\footnote{https://github.com/MCC-WH/SSP}\footnote{https://github.com/yanbai1993/Dual-Tuning/tree/master} released by the authors. For RM, RBCL, and CoReS, we directly use the experimental results reported by the authors. In particular, SSPL~\cite{SSPL} is specifically designed for the asymmetric retrieval task. It typically assumes the existence of a pretrained large gallery model and trains a small yet compatible query model by transferring the knowledge from the gallery model to the query model. Directly applying SSPL to BCL makes it difficult to obtain a new model that outperforms the old model. Hence, we introduce a cross-entropy loss for SSPL, as many BCL algorithms do, which is referred to as SSPL-CE.
In these experiments, ResNet18 (R18), ResNet50 (R50), and ResNet100 (R100) are used as the backbones.

\begin{table}[t!]
\centering
\setlength{\tabcolsep}{6pt}
\renewcommand{\arraystretch}{1.1}
\caption{Results of different algorithms on In-shop in the single-step BCL experiment. $\dagger$ denotes that the model is trained independently.}
\resizebox{0.95\linewidth}{!}{
\begin{tabular}{llcccccc}
\toprule
\multirow{2}{*}{\shortstack{Allocation\\Type}} &\multirow{2}{*}{Methods} &\multicolumn{2}{c}{mAP} &\multirow{2}{*}{$\mathcal{P}_{up}$} &\multirow{2}{*}{$\mathcal{P}_{comp}$} &\multirow{2}{*}{$\mathcal{P}_{1-score}$}\\ 
\cline{3-4}  
                     &  &self-test         & cross-test           \\ 
\midrule 
            \multirow{12}{*}{\shortstack{Data-\\Extension\\(30\%$\rightarrow$100\%)}} 
            & \multicolumn{1}{>{\cellcolor[gray]{0.9}}l}{$\dagger$R18 (old)} 
            & \multicolumn{1}{>{\cellcolor[gray]{0.9}}c}{53.26} 
            & \multicolumn{1}{>{\cellcolor[gray]{0.9}}c}{--} 
            & \multicolumn{1}{>{\cellcolor[gray]{0.9}}c}{--} 
            & \multicolumn{1}{>{\cellcolor[gray]{0.9}}c}{--} 
            & \multicolumn{1}{>{\cellcolor[gray]{0.9}}c}{--} \\
            & \multicolumn{1}{>{\cellcolor[gray]{0.9}}l}{$\dagger$R18 (new)} 
            & \multicolumn{1}{>{\cellcolor[gray]{0.9}}c}{71.24} 
            & \multicolumn{1}{>{\cellcolor[gray]{0.9}}c}{--} 
            & \multicolumn{1}{>{\cellcolor[gray]{0.9}}c}{--} 
            & \multicolumn{1}{>{\cellcolor[gray]{0.9}}c}{--} 
            & \multicolumn{1}{>{\cellcolor[gray]{0.9}}c}{--} \\
           &BCT~\cite{BCT}   &65.30       & 54.36     &47.92&51.53&49.66  \\
             &UniBCT~\cite{unibct}  & 61.83       & 54.07  &46.70&51.13&48.81\\ 
             &RACT~\cite{hot-refresh}  &64.22       &54.65 &47.34&51.86&49.50 \\ 
             &AdvBCT~\cite{advbct}  &65.45       & 54.15  &47.97&51.24&49.55\\  
             & Dual-Tuning~\cite{DualTuning}&64.02 & 55.06 &47.47&52.50&49.86 \\
             & SSPL-CE~\cite{SSPL} &60.74&55.02&46.32&52.45&49.19\\
             & SSPL~\cite{SSPL} &52.86&52.69&43.59&49.21&46.23\\
             & BT$^2$~\cite{bt2}& 62.33&55.01&46.88&52.43&49.50\\ 
             &\textbf{NDPP} (Ours) &  \textbf{67.55}       & \underline{56.32} &\textbf{48.71}&\underline{54.24}&\textbf{51.33} \\  
             &\textbf{ODPP} (Ours) &  \underline{65.74}       & \textbf{56.50} &\underline{48.07}&\textbf{54.49}&\underline{51.08} \\        
\midrule
            \multirow{12}{*}{\shortstack{Backbone-\\Extension\\(R18$\rightarrow$R50)}} 
            & \multicolumn{1}{>{\cellcolor[gray]{0.9}}l}{$\dagger$R18 (old)}   
            & \multicolumn{1}{>{\cellcolor[gray]{0.9}}c}{53.26}  
            & \multicolumn{1}{>{\cellcolor[gray]{0.9}}c}{--}  
            & \multicolumn{1}{>{\cellcolor[gray]{0.9}}c}{--}  
            & \multicolumn{1}{>{\cellcolor[gray]{0.9}}c}{--}  
            & \multicolumn{1}{>{\cellcolor[gray]{0.9}}c}{--} \\
            & \multicolumn{1}{>{\cellcolor[gray]{0.9}}l}{$\dagger$R50 (new)}  
            & \multicolumn{1}{>{\cellcolor[gray]{0.9}}c}{71.89}  
            & \multicolumn{1}{>{\cellcolor[gray]{0.9}}c}{--}  
            & \multicolumn{1}{>{\cellcolor[gray]{0.9}}c}{--}  
            & \multicolumn{1}{>{\cellcolor[gray]{0.9}}c}{--}  
            & \multicolumn{1}{>{\cellcolor[gray]{0.9}}c}{--} \\
           &BCT~\cite{BCT} &63.61       & 51.37      &47.12&47.47&47.29 \\
             &UniBCT~\cite{unibct} & 64.48       & 55.66 &47.43&53.22&50.15 \\ 
             &RACT~\cite{hot-refresh} & 65.22       & 54.63  &47.89&51.90&49.82\\              
             &AdvBCT~\cite{advbct} &63.10       & 54.45 &46.95&51.60&49.16 \\ 
            &Dual-Tuning~\cite{DualTuning}  & 63.91&53.93&47.23&50.90&48.99\\
            & SSPL-CE~\cite{SSPL}  & 60.33& 54.20&45.99&51.26&48.48\\
             & SSPL~\cite{SSPL} & 53.01& 52.87&43.47&49.48&46.28\\
            &BT$^2$~\cite{bt2} &64.30 & 54.37&47.36&51.49&49.34\\ 
             &\textbf{NDPP} (Ours) & \textbf{71.03}       & \underline{56.33} &\textbf{49.70}&\underline{54.11}&\textbf{51.81} \\ 
             &\textbf{ODPP} (Ours) & \underline{66.74}      & \textbf{56.54}&\underline{48.21}&\textbf{54.39}&\underline{51.11} \\
\bottomrule
\end{tabular}}
\label{Tab:inshop}
\vspace{-2mm}
\end{table}

\begin{table}[t!]
\centering
\setlength{\tabcolsep}{6pt}
\renewcommand{\arraystretch}{1.1}
\caption{Results of different algorithms on Market-1501 in the single-step BCL experiment. $\dagger$ denotes that the model is trained independently. $\ddagger$ denotes that the results of the model are reported by the authors, which are presented solely for reference, as their independent old model shows slight performance differences from ours.}
\resizebox{0.95\linewidth}{!}{
\begin{tabular}{llccccc}
\toprule
\multirow{2}{*}{\shortstack{Allocation\\Type}} &\multirow{2}{*}{Methods} & \multicolumn{2}{c}{Recall@1} && \multicolumn{2}{c}{mAP}\\ 
\cline{3-4} \cline{6-7}  
                     &  & self-test         & cross-test          && self-test         & cross-test           \\ 
\midrule 
            \multirow{12}{*}{\shortstack{Data-\\Extension\\(50\%$\rightarrow$100\%)}} 
            & \multicolumn{1}{>{\cellcolor[gray]{0.9}}l}{$\dagger$R18 (old)} 
            & \multicolumn{1}{>{\cellcolor[gray]{0.9}}c}{82.19} 
            & \multicolumn{1}{>{\cellcolor[gray]{0.9}}c}{--} 
            & \multicolumn{1}{>{\cellcolor[gray]{0.9}}c}{} 
            & \multicolumn{1}{>{\cellcolor[gray]{0.9}}c}{62.09} 
            & \multicolumn{1}{>{\cellcolor[gray]{0.9}}c}{--}  \\
            & \multicolumn{1}{>{\cellcolor[gray]{0.9}}l}{$\dagger$R18 (new)} 
            & \multicolumn{1}{>{\cellcolor[gray]{0.9}}c}{92.10} 
            & \multicolumn{1}{>{\cellcolor[gray]{0.9}}c}{--} 
            & \multicolumn{1}{>{\cellcolor[gray]{0.9}}c}{} 
            & \multicolumn{1}{>{\cellcolor[gray]{0.9}}c}{79.69} 
            & \multicolumn{1}{>{\cellcolor[gray]{0.9}}c}{--}  \\
            & BCT~\cite{BCT} & 91.24 &85.93 &&77.41& 67.28\\
            &UniBCT~\cite{unibct} & 89.51&84.53&&72.97&64.19\\
            & RACT~\cite{hot-refresh} & 90.35 &83.74 && 74.18 &63.51\\
            & AdvBCT~\cite{advbct} & 91.56 & 86.16 &&78.97 & 68.05\\
            & $\ddagger$Dual-Tuning~\cite{DualTuning} &92.04 & 86.67 && 80.75 &69.53 \\
           & SSPL-CE~\cite{SSPL} &91.53&86.25&&79.69&68.41\\
            & SSPL~\cite{SSPL} & 81.87&81.94&& 61.77&61.86\\
            &BT$^2$~\cite{bt2}&90.91&85.42&&77.64&67.88\\
            & $\ddagger$RM~\cite{RM} &--& 87.00 &&--&69.20 \\
            &\textbf{NDPP} (Ours) & \textbf{92.84}       &\underline{87.86}     && \textbf{82.02}       & \textbf{70.55}
             \\  
            &\textbf{ODPP} (Ours) & \underline{92.81}      & \textbf{88.00}      && \underline{81.90} & \underline{70.15}    \\ 
\midrule 
            \multirow{12}{*}{\shortstack{Data-\\Extension\\(10\%$\rightarrow$100\%)}} 
            & \multicolumn{1}{>{\cellcolor[gray]{0.9}}l}{$\dagger$R50 (old)} 
            & \multicolumn{1}{>{\cellcolor[gray]{0.9}}c}{63.27} 
            & \multicolumn{1}{>{\cellcolor[gray]{0.9}}c}{--} 
            & \multicolumn{1}{>{\cellcolor[gray]{0.9}}c}{} 
            & \multicolumn{1}{>{\cellcolor[gray]{0.9}}c}{37.80} 
            & \multicolumn{1}{>{\cellcolor[gray]{0.9}}c}{--}  \\
            & \multicolumn{1}{>{\cellcolor[gray]{0.9}}l}{$\dagger$R50 (new)} 
            & \multicolumn{1}{>{\cellcolor[gray]{0.9}}c}{93.97} 
            & \multicolumn{1}{>{\cellcolor[gray]{0.9}}c}{--} 
            & \multicolumn{1}{>{\cellcolor[gray]{0.9}}c}{} 
            & \multicolumn{1}{>{\cellcolor[gray]{0.9}}c}{84.42} 
            & \multicolumn{1}{>{\cellcolor[gray]{0.9}}c}{--}  \\
            & BCT~\cite{BCT} &90.88&74.05&&76.88&49.27\\
            & UniBCT~\cite{unibct}&81.38&69.80&&59.14&43.72 \\
            & RACT~\cite{hot-refresh} &91.54&77.35&&78.43&52.83\\
            & AdvBCT~\cite{advbct} &93.26 &78.36&&82.76&55.35\\
            & Dual-Tuning~\cite{DualTuning} &92.58&82.54&&81.09&59.37  \\
            & SSPL-CE~\cite{SSPL} &90.24&82.49&&79.69&56.98\\
            & SSPL~\cite{SSPL} & 63.96&63.89&&38.15&37.90\\
            & BT$^2$~\cite{bt2} &85.18&75.45 &&66.48&50.65\\                  
            & $\ddagger$RBCL~\cite{RBCL} &93.70&82.70  &&85.00&61.30 \\
            &\textbf{NDPP} (Ours) &  \textbf{94.45}     &\textbf{84.26}     &&  \textbf{86.05}     & \textbf{62.52}\\  
            &\textbf{ODPP} (Ours) &  \underline{94.30}    &  \underline{83.49}    && \underline{85.59} &\underline{62.40}    \\   
\bottomrule
\end{tabular}}
\label{Tab:market1501}
\vspace{-3mm}
\end{table}

\vspace{2mm}
\subsubsection{Single-step backward-compatible learning} We conduct the single-step BCL experiment on the landmark, commodity, and person Re-ID benchmarks.

\vspace{2mm}
\noindent\textbf{Landmark.} Table~\ref{Tab:gldv2} presents the experimental results. The models are trained on the GLDv2 training set. Following AdvBCT~\cite{advbct}, we report the average of the $\mathcal{P}_{comp}$, $\mathcal{P}_{up}$ and $\mathcal{P}_{1-score}$ on RParis, ROxford, and GLDv2-test.
The old model and new model are trained with 9\% and 30\% of classes from GLDv2, respectively. For the data-extension ($9\%\rightarrow 30\%$) experiment, where ResNet18 is used as the backbone, NDPP and ODPP perform favorably against state-of-the-art BCL methods on self-test and cross-test evaluations. Although BT$^2$~\cite{bt2} leverages extra dimensions to reconcile backward compatibility with new model performance, NDPP and ODPP outperform it with substantial gains on all datasets. This demonstrates the effectiveness of our prototype perturbation mechanism. Notably, SSPL~\cite{SSPL} fails to achieve backward compatibility, while SSPL-CE successfully achieves backward compatibility. Specifically, NDPP outperforms ODPP slightly in this experimental setting on the overall metric $\mathcal{P}_{1-score}$. Only 9\% of the GLDv2 dataset (7,318 classes) is utilized for training the old model. These classes remain relatively separable in the old feature space. Consequently, perturbations computed based on local information, \ie~calculated by NDPP, can produce satisfactory results.

For the backbone-extension (R18$\rightarrow$R50) experiment, our NDPP and ODPP also outperform state-of-the-art BCL algorithms on most metrics. These performance gains indicate that our prototype perturbation can also facilitate model upgrades by altering the backbone. The backbone-extension scenario presents greater challenges for perturbation calculations involving both old and new prototypes. This is because prototypes extracted from models with different architectures exhibit more pronounced discrepant distributions than those derived from identical architectures. As a result, computing appropriate perturbations for the old prototypes based on the old (R18) and new (R50) prototypes becomes increasingly complex, especially when the number of classes used to train the new model reaches 24,393. Under such a condition, ODPP, with enhanced ability to learn appropriate perturbations in such challenging circumstances, demonstrates a slight overall performance advantage over NDPP.

\vspace{2mm}
\noindent\textbf{Commodity.} We conduct experiments on In-shop using the same data allocation and backbone setting as GLDv2. As shown in Table~\ref{Tab:inshop}, NDPP and ODPP perform favorably against state-of-the-art BCL algorithms in both the data-extension and backbone-extension experiments. Most existing methods, including AdvBCT, Dual-Tuning, SSPL-CE, and BT2, achieve cross-test performance comparable to our approaches. However, their self-test performance is substantially lower than that of our NDPP and ODPP. These results demonstrate that prototype perturbation effectively relaxes the backward-compatible constraint and enhances the discriminative ability of the new model. Moreover, NDPP demonstrates superior overall performance compared to ODPP in both the data-extension and backbone-extension experiments, which differs from the results observed on GLDv2. We attribute this discrepancy to the small scale of In-Shop, where the old and new models are trained on 1,199 and 3,997 classes, respectively. This smaller scale simplifies the computation of perturbations, enabling NDPP to identify appropriate perturbations based solely on local neighborhoods. In contrast, the perturbations learned by ODPP may overfit the sparse feature distribution of the old model during BCL, resulting in slightly inferior performance compared with NDPP.

\begin{table*}[t]
\centering
\setlength{\tabcolsep}{9pt}
\renewcommand{\arraystretch}{1.1}
\caption{Detailed results of different algorithms of the data-extension setting in the sequential BCL experiment. $\phi_1$, $\phi_2$, and $\phi_3$ are trained with 9\%, 30\%, and 100\% of data, respectively. ResNet18 is used as the backbone in $\phi_1$, $\phi_2$, and $\phi_3$. $\mathcal{P}_{up}$, $\mathcal{P}_{comp}$, and $\mathcal{P}_{1-score}$ are calculated based on the mAP over RParis, ROxford, and GLDv2-test.}
\resizebox{0.95\linewidth}{!}{
\begin{tabular}{lccccccccccccccc}
\toprule
\multirow{2}{*}{Method} &\multirow{2}{*}{\shortstack{9\%\\Data}}& \multirow{2}{*}{\shortstack{30\%\\Data}}& \multirow{2}{*}{\shortstack{100\%\\Data}}& \multicolumn{2}{c}{RParis (mAP)} && \multicolumn{2}{c}{ROxford (mAP)}&& \multicolumn{2}{c}{GLDv2-test (mAP)} &\multirow{2}{*}{$\mathcal{P}_{up}$} &\multirow{2}{*}{$\mathcal{P}_{comp}$} &\multirow{2}{*}{$\mathcal{P}_{1-score}$}\\ 
\cline{5-6} \cline{8-9} \cline{11-12}
                       &&&& self-test         & cross-test        && self-test         & cross-test && self-test &cross-test          \\ 
\midrule
              \multirow{3}{*}{\shortstack[l]{Independent\\Training}}   & $\phi_1$     & --         &--& 67.31 &--  && 41.82  &-- &&7.30 &-- &--&--&-- \\
               &  --    & $\phi_2$         &--& 75.08 &--  && 55.77  &-- &&12.08 &--  &--&--&--\\
              & --     & --         &$\phi_3$&81.13  &--  &&62.38   &-- &&15.83 &-- &--&--&-- \\

\midrule

              BCT   & $\phi_1$     & $\phi_2$         &--& 71.52 &68.37  && 51.16  &44.75 &&10.38 &\underline{8.72} &47.75  &55.34    &51.23  \\
              AdvBCT   & $\phi_1$     & $\phi_2$         &--& 74.24 &68.70  && 54.08  &42.88 &&11.57 &7.94 &49.30&53.23&51.19  \\
              NDPP (Ours)  & $\phi_1$     & $\phi_2$         &--& \underline{76.21} &\textbf{71.10}  && \textbf{58.02}  &\textbf{46.42} &&\textbf{12.67} &\textbf{8.88} &\textbf{50.87}&\textbf{59.44}&\textbf{54.80}\\
              ODPP (Ours)   & $\phi_1$     & $\phi_2$         &--& \textbf{76.84} &\underline{70.99}  && \underline{56.07}  &\underline{44.81} &&\underline{12.32} &8.56 &\underline{50.55}&\underline{58.75}&\underline{54.32}\\
\midrule
              
               BCT & $\phi_1$     &  --        &$\phi_3$ & 78.11 &70.24  && 59.38  &46.68 &&14.15 &8.67  &48.41  &55.06   &51.52\\
               AdvBCT & $\phi_1$     & --        &$\phi_3$& \underline{81.06} &68.57  && \underline{62.71}  &45.95 &&16.07 &8.67 &50.16  &53.36   &51.71\\
               NDPP (Ours) & $\phi_1$     &    --     &$\phi_3$& 80.61 &\underline{70.69}  && 61.62  &\underline{46.88} &&\textbf{16.85} &\underline{9.40} &\underline{50.38}  &\underline{56.11}   &\underline{53.09}\\
              ODPP (Ours)& $\phi_1$     &  --       &$\phi_3$& \textbf{81.29} &\textbf{71.87}  && \textbf{64.46}  &\textbf{47.85} &&\underline{16.58} &\textbf{9.48} &\textbf{50.69}  &\textbf{57.27}   &\textbf{53.77}\\
\midrule
              
              BCT &  --    &    $\phi_2$      &$\phi_3$& 78.11 &74.21  && 59.38  &52.34 &&14.15 &11.51 &48.41  &43.31   &45.59\\
              AdvBCT  & --     &$\phi_2$          &$\phi_3$& \underline{81.06} &75.16  && \underline{62.71}  &54.42 &&16.07 &11.66 &50.16  &47.48   &48.76\\
              NDPP (Ours)&   --   &  $\phi_2$        &$\phi_3$& 80.61 &\underline{76.25}  && 61.62  &\underline{54.50} &&\textbf{16.85} &\textbf{13.71} &\underline{50.38}  &\underline{53.58}   &\underline{51.78}\\
              ODPP (Ours)&  --    &   $\phi_2$       &$\phi_3$& \textbf{81.29} &\textbf{77.56}  && \textbf{64.46}  &\textbf{55.05} &&\underline{16.58} &\underline{13.48} &\textbf{50.69}  &\textbf{55.54}   &\textbf{52.84}\\              
\bottomrule
\end{tabular}}
\label{Tab:sequential_BCL}
\vspace{-3mm}
\end{table*}

\begin{table}[t!]
\centering
\setlength{\tabcolsep}{12pt}
\renewcommand{\arraystretch}{1.1}
\caption{Compatibility evaluation for person re-identification on the Market1501 dataset for one and two-step upgrading. We report $AC$ and $AM$ calculated based on mAP. $\ddagger$ denotes that the result is reported in~\cite{Cores}.}
\resizebox{0.9\linewidth}{!}{
\begin{tabular}{lccccc}
\toprule
 \multirow{2}{*}{Methods} & \multicolumn{2}{c}{One-upgrade}&& \multicolumn{2}{c}{Two-upgrades}\\
 \cline{2-3} \cline{5-6}
 & $AC$ & $AM$ && $AC$ & $AM$\\
\midrule
$\ddagger$LwF~\cite{lwf} & 0 & 0.300 && 0 & 0.210 \\
$\ddagger$BCT~\cite{BCT} &\textbf{1}& 0.490&&0.3&0.500\\
$\ddagger$CoReS~\cite{Cores} & \textbf{1}& 0.570 && \textbf{1} & 0.510\\
\textbf{NDPP} (ours)&\textbf{1} &\textbf{0.606} &&\textbf{1}&\textbf{0.652} \\
\textbf{ODPP} (ours)&\textbf{1} &\underline{0.593} &&\textbf{1}&\underline{0.644} \\
\bottomrule
\end{tabular}}
\label{Tab:Cores_market1501}
\vspace{-3mm}
\end{table}

\begin{figure*}[t!]
\centering
\includegraphics[width=0.7\linewidth]{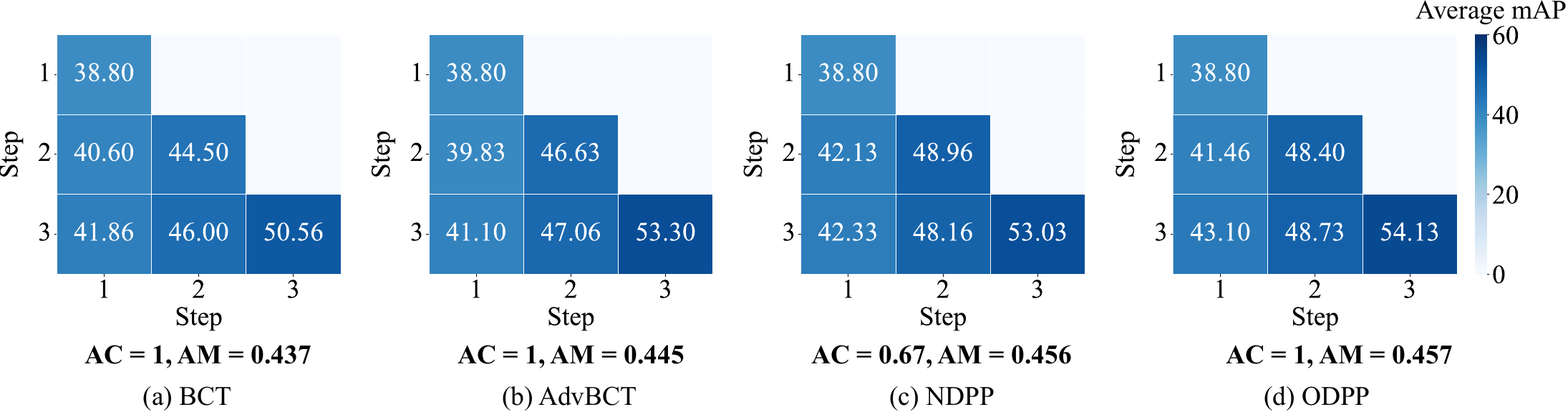}
\vspace{-1mm}
\caption{Compatibility matrices of 4 methods on GLDv2 in the 2-step model upgrade experiment. Models are sequentially learned with 9\%, 30\%, and 100\% of training data. We report the average mAP on RParis, ROxford, and GLDv2-test. We also report $AC$ and $AM$ calculated based on mAP.}
\label{Fig:gldv2_average_upgrades}
\vspace{-3mm}
\end{figure*}

\vspace{2mm}
\noindent\textbf{Person Re-ID.} Table~\ref{Tab:market1501} presents the results on Market-1501. Herein, we conduct the data-extension experiment with two different data allocation settings, following RM~\cite{RM} and RBCL~\cite{RBCL}, respectively. In the first data allocation setting ($50\% \rightarrow 100\%$), where ResNet18 is used as the backbone, Dual-Tuning and RM achieve impressive cross-test performance in both Recall@1 and mAP. Compared with them, our NDPP and ODPP further improve the cross-test performance. This indicates that our prototype perturbation mechanism does not compromise the backward compatibility of the model. Moreover, it can further enhance the discriminative ability of the new model, evidenced by the superior self-test performance compared to Dual-Tuning and even the independently trained new model. In the second data allocation setting ($10\% \rightarrow 100\%$), where ResNet50 is used as the backbone, our NDPP and ODPP achieve substantial performance gains of $24.72\%$ and $24.60\%$ in cross-test mAP, compared with the self-test mAP of the old model. In both data allocation settings, NDPP demonstrates slightly better performance than the ODPP, which we attribute to the relatively small scale of Market-1501 (containing only 751 classes in total). In this condition, NDPP can compute appropriate perturbations based only on neighboring prototypes.

\vspace{2mm}
\subsubsection{Sequential backward-compatible learning}
We first conduct the sequential BCL experiment on the large-scale dataset GLDv2. Specifically, we train three models using 9\%, 30\%, and 100\% of the data sequentially, which are denoted by $\phi_1$, $\phi_2$, and $\phi_3$, respectively. $\phi_2$ is supervised to be compatible with $\phi_1$, and $\phi_3$ is further restrained to be compatible with $\phi_2$. A pretrained ResNet18 is used as the backbone. As shown in Table~\ref{Tab:sequential_BCL}, NDPP and ODPP perform better than BCT and AdvBCT. The comparison demonstrates the effectiveness of our approaches in the sequential model upgrade scenario. Following~\cite{Cores,CL2R,biondi2024stationary,LCE}, we also reorganize the experimental results into the compatibility matrix for a more intuitive comparison, as shown in Figure~\ref{Fig:gldv2_average_upgrades}. We can observe that ODPP obtains the best self-test average mAP of 54.13 and the best cross-test average mAP of 48.73 (with $\phi_2$) and 43.10 (with $\phi_1$) after sequential model upgrade. NDPP achieves a slightly lower self-test average mAP compared to AdvBCT, but demonstrates superior cross-test average mAP when evaluated with previous models. Moreover, NDPP performs better than ODPP on the first BCL step (9\% data$\rightarrow$30\% data), while ODPP obtains better performance on the second BCL step (30\% data $\rightarrow$100\% data) in both Table~\ref{Tab:sequential_BCL} and Figure~\ref{Fig:gldv2_average_upgrades}. This difference can be attributed to their underlying mechanisms: ODPP iteratively refines perturbations by leveraging the overall feature distributions, enabling it to generate more appropriate perturbations when the training dataset contains a large number of classes (24,393 and 81,313 classes for the old and new models, respectively, in the second BCL step). In contrast, NDPP computes perturbations based solely on neighboring prototypes, which is particularly effective when the number of classes is relatively small (7,318 and 24,393 classes for the old and new models, respectively).

Besides, we also conduct the sequential BCL experiment on Market-1501, following the setting of CoReS~\cite{Cores}. Specifically, 33.3\% of classes are used to train the initial model, 66.7\% of classes are employed for the first upgrade, and the entire dataset is utilized for the second upgrade. A pretrained ResNet101 is used as the backbone, and the new model is trained for 25 epochs for each model upgrade, which is consistent with~\cite{Cores}. Table~\ref{Tab:Cores_market1501} presents the \textit{AC} and \textit{AM} of different methods, which are calculated based on mAP. CoReS~\cite{Cores} maintains complete backward compatibility, whereas BCT loses this capability after two upgrade steps. Both our NDPP and ODPP achieve full backward compatibility within two upgrade steps and perform favorably against BCT and CoReS in \textit{AM}, further demonstrating the effectiveness of the prototype perturbation mechanism in sequential model upgrade.

\begin{table}[t!]
\centering
\setlength{\tabcolsep}{7pt}
\renewcommand{\arraystretch}{1.1}
\caption{Results of the single-step BCL experiment on the text-based person retrieval datasets RSTPReid~\cite{RSTPReid}. Here, we use the pretrained APTM~\cite{APTM} as the embedding model. APTM (old) and APTM (new) are trained independently with 50\% and 100\% of data, respectively.}
\vspace{-2mm}
\resizebox{0.95\linewidth}{!}{
\begin{tabular}{llccccc}
\toprule
\multirow{2}{*}{\shortstack{Allocation\\Type}} &\multirow{2}{*}{Methods} & \multicolumn{2}{c}{Recall@1} && \multicolumn{2}{c}{mAP}\\ 
\cline{3-4} \cline{6-7}  
                     &  & self-test         & cross-test          && self-test         & cross-test           \\ 
\midrule 
            \multirow{4}{*}{\shortstack{Data-\\Extension\\(50\%$\rightarrow$100\%)}} 
            & \multicolumn{1}{>{\cellcolor[gray]{0.9}}l}{$\dagger$APTM (old)} 
            & \multicolumn{1}{>{\cellcolor[gray]{0.9}}c}{57.85} 
            & \multicolumn{1}{>{\cellcolor[gray]{0.9}}c}{--} 
            & \multicolumn{1}{>{\cellcolor[gray]{0.9}}c}{} 
            & \multicolumn{1}{>{\cellcolor[gray]{0.9}}c}{45.55} 
            & \multicolumn{1}{>{\cellcolor[gray]{0.9}}c}{--}  \\
            & \multicolumn{1}{>{\cellcolor[gray]{0.9}}l}{$\dagger$APTM (new)} 
            & \multicolumn{1}{>{\cellcolor[gray]{0.9}}c}{62.60} 
            & \multicolumn{1}{>{\cellcolor[gray]{0.9}}c}{--} 
            & \multicolumn{1}{>{\cellcolor[gray]{0.9}}c}{} 
            & \multicolumn{1}{>{\cellcolor[gray]{0.9}}c}{49.78} 
            & \multicolumn{1}{>{\cellcolor[gray]{0.9}}c}{--}  \\
            &\textbf{NDPP} (Ours) & \textbf{63.80}       &\textbf{61.20}     && \textbf{51.19}       & \textbf{47.64}
             \\  
            &\textbf{ODPP} (Ours) & \underline{62.75}      &  \underline{61.15}      && \underline{50.49} & \underline{47.54}    \\  
\bottomrule
\end{tabular}}
\label{Tab:RSTPReid}
\vspace{-3mm}
\end{table}

\begin{table*}[h!]
\centering
\setlength{\tabcolsep}{10pt}
\renewcommand{\arraystretch}{1.1}
\caption{Experimental results of five variants of our method on RParis, ROxford, and GLDv2-test. $\dagger$ denotes that the model is trained independently. $\mathcal{P}_{up}$, $\mathcal{P}_{comp}$, and $\mathcal{P}_{1-score}$ are calculated based on the mAP on RParis, ROxford, and GLDv2-test.}
\resizebox{0.95\linewidth}{!}{
\begin{tabular}{lcccccccccccc}
\toprule
\multirow{2}{*}{Variants} & \multicolumn{2}{c}{RParis} && \multicolumn{2}{c}{ROxford} && \multicolumn{2}{c}{GLDv2-test} &\multirow{2}{*}{$\mathcal{P}_{up}$} &\multirow{2}{*}{$\mathcal{P}_{comp}$} &\multirow{2}{*}{$\mathcal{P}_{1-score}$}\\ 
\cline{2-3} \cline{5-6} \cline{8-9} 
                       & self-test  & cross-test && self-test  & cross-test && self-test  & cross-test  \\
\midrule
        \rowcolor[gray]{0.9} $\dagger$R18 (old)  & 67.31       & --   &&41.82       &--   &&7.30  &--   & -- & -- & -- \\
        \rowcolor[gray]{0.9} $\dagger$R18 (new)  & 75.08       & --   &&55.77       &--  &&12.08  &--   & -- & -- & --   \\
        Baseline & 75.42 & 69.87  && 54.97   & 44.31 && 11.81 & 8.58 & 49.73 & 56.42 & 52.86\\
        NDPP-old & 76.07 & 70.15 && 53.48 & 44.35 && \underline{12.56} & \underline{8.73}&50.10&56.99 & 53.32  \\
        NDPP     & \underline{76.21}      &  \textbf{71.10}      && \textbf{58.02}       & \textbf{46.42} &&\textbf{12.67}  &\textbf{8.88}  &\textbf{50.87}&\textbf{59.44}& \textbf{54.80} \\     
        ODPP-old & 75.97 & 70.39 && 54.46 & 45.43 && 12.26 & 8.64 &49.98&57.59& 53.51\\
        ODPP     & \textbf{76.84}       & \underline{70.99}      && \underline{56.45}       & \underline{46.24} && 12.44 & 8.60 &\underline{50.55}&\underline{58.75} &\underline{54.32} \\
\bottomrule
\end{tabular}}
\label{Tab:ablation}
\vspace{-3mm}
\end{table*}

\subsection{Extension to Multimodal Retrieval}
Prototype perturbation can also be applied to facilitate backward-compatible upgrades of multimodal retrieval models. To validate our approach, we conduct the BCL experiment on RSTPREID~\cite{RSTPReid}, a text-based person retrieval dataset, where $50\%$ and $100\%$ of data are used to train the old and new models, respectively. The pretrained APTM~\cite{APTM} is used as the embedding model. Technically, we employ the contrastive loss with prototype perturbation to constrain the compatibility between the image embeddings of the new and old models. With the multimodal alignment constraints inherent in APTM, both the image and text embeddings of the new model can be compatible with the existing image and text embeddings. We follow the training details of~\cite{APTM} in this experiment. Table~\ref{Tab:RSTPReid} presents the results of our NDPP and ODPP. The cross-test mAP of NDPP and ODPP has improved by $2.09\%$ and $2.19\%$, respectively, compared to the self-test mAP of the old model. This improvement suggests that our prototype perturbation can be effectively extended to multimodal data.

\begin{table*}[h!]
\centering
\setlength{\tabcolsep}{9pt}
\renewcommand{\arraystretch}{1.1}
\caption{Results on RParis, ROxford, and GLDv2-test in the single-step BCL experiment using the pretrained or randomly initialized backbone. $\dagger$ denotes that the model is trained independently. $\mathcal{P}_{up}$, $\mathcal{P}_{comp}$, and $\mathcal{P}_{1-score}$ are calculated based on the mAP over RParis, ROxford, and GLDv2-test.}
\vspace{-2mm}
\resizebox{0.95\linewidth}{!}{
\begin{tabular}{clcccccccccccc}
\toprule
 \multirow{2}{*}{\shortstack{Pretrained\\Backbone}}  &\multirow{2}{*}{Methods} & \multicolumn{2}{c}{RParis (mAP)} && \multicolumn{2}{c}{ROxford (mAP)} && \multicolumn{2}{c}{GLDv2-test (mAP)} &\multirow{2}{*}{$\mathcal{P}_{up}$} &\multirow{2}{*}{$\mathcal{P}_{comp}$} &\multirow{2}{*}{$\mathcal{P}_{1-score}$}\\ 
\cline{3-4} \cline{6-7} \cline{9-10} 
                     & &self-test         & cross-test          && self-test         & cross-test  && self-test         & cross-test         \\ 
\midrule
            \multirow{5}{*}{\ding{52}} 
            & \multicolumn{1}{>{\cellcolor[gray]{0.9}}l}{$\dagger$R18 (old)}  
            & \multicolumn{1}{>{\cellcolor[gray]{0.9}}c}{67.31}  
            & \multicolumn{1}{>{\cellcolor[gray]{0.9}}c}{--}  
            & \multicolumn{1}{>{\cellcolor[gray]{0.9}}c}{}  
            & \multicolumn{1}{>{\cellcolor[gray]{0.9}}c}{41.82}  
            & \multicolumn{1}{>{\cellcolor[gray]{0.9}}c}{--}  
            & \multicolumn{1}{>{\cellcolor[gray]{0.9}}c}{}  
            & \multicolumn{1}{>{\cellcolor[gray]{0.9}}c}{7.30}  
            & \multicolumn{1}{>{\cellcolor[gray]{0.9}}c}{--}  
            & \multicolumn{1}{>{\cellcolor[gray]{0.9}}c}{--}  
            & \multicolumn{1}{>{\cellcolor[gray]{0.9}}c}{--}  
            & \multicolumn{1}{>{\cellcolor[gray]{0.9}}c}{--} \\
            & \multicolumn{1}{>{\cellcolor[gray]{0.9}}l}{$\dagger$R18 (new)}  
            & \multicolumn{1}{>{\cellcolor[gray]{0.9}}c}{75.08}  
            & \multicolumn{1}{>{\cellcolor[gray]{0.9}}c}{--}  
            & \multicolumn{1}{>{\cellcolor[gray]{0.9}}c}{}  
            & \multicolumn{1}{>{\cellcolor[gray]{0.9}}c}{55.77}  
            & \multicolumn{1}{>{\cellcolor[gray]{0.9}}c}{--}  
            & \multicolumn{1}{>{\cellcolor[gray]{0.9}}c}{}  
            & \multicolumn{1}{>{\cellcolor[gray]{0.9}}c}{12.08}  
            & \multicolumn{1}{>{\cellcolor[gray]{0.9}}c}{--}  
            & \multicolumn{1}{>{\cellcolor[gray]{0.9}}c}{--}  
            & \multicolumn{1}{>{\cellcolor[gray]{0.9}}c}{--}  
            & \multicolumn{1}{>{\cellcolor[gray]{0.9}}c}{--} \\
            &Baseline & 75.42 & 69.87  && 54.97   & 44.31 && 11.81 & 8.58 & 49.73 & 56.42 & 52.86\\
            & \textbf{NDPP} (Ours)  & \underline{76.21}      &  \textbf{71.10}      && \textbf{58.02}       & \textbf{46.42} &&\textbf{12.67}  &\textbf{8.88}  & \textbf{50.87} & \textbf{59.44} & \textbf{54.80} \\     
            & \textbf{ODPP} (Ours)  & \textbf{76.84}       & \underline{70.99}      && \underline{56.45}       & \underline{46.24} && 12.44 & 8.60 & \underline{50.55} & \underline{58.75} & \underline{54.32} \\
\midrule
            \multirow{5}{*}{\ding{56}}
            & \multicolumn{1}{>{\cellcolor[gray]{0.9}}l}{$\dagger$R18 (old)}  
            & \multicolumn{1}{>{\cellcolor[gray]{0.9}}c}{51.89}  
            & \multicolumn{1}{>{\cellcolor[gray]{0.9}}c}{--}  
            & \multicolumn{1}{>{\cellcolor[gray]{0.9}}c}{}  
            & \multicolumn{1}{>{\cellcolor[gray]{0.9}}c}{29.63}  
            & \multicolumn{1}{>{\cellcolor[gray]{0.9}}c}{--}  
            & \multicolumn{1}{>{\cellcolor[gray]{0.9}}c}{}  
            & \multicolumn{1}{>{\cellcolor[gray]{0.9}}c}{3.35}  
            & \multicolumn{1}{>{\cellcolor[gray]{0.9}}c}{--}  
            & \multicolumn{1}{>{\cellcolor[gray]{0.9}}c}{--}  
            & \multicolumn{1}{>{\cellcolor[gray]{0.9}}c}{--}  
            & \multicolumn{1}{>{\cellcolor[gray]{0.9}}c}{--} \\
            & \multicolumn{1}{>{\cellcolor[gray]{0.9}}l}{$\dagger$R18 (new)}  
            & \multicolumn{1}{>{\cellcolor[gray]{0.9}}c}{68.71}  
            & \multicolumn{1}{>{\cellcolor[gray]{0.9}}c}{--}  
            & \multicolumn{1}{>{\cellcolor[gray]{0.9}}c}{}  
            & \multicolumn{1}{>{\cellcolor[gray]{0.9}}c}{49.48}  
            & \multicolumn{1}{>{\cellcolor[gray]{0.9}}c}{--}  
            & \multicolumn{1}{>{\cellcolor[gray]{0.9}}c}{}  
            & \multicolumn{1}{>{\cellcolor[gray]{0.9}}c}{9.74}  
            & \multicolumn{1}{>{\cellcolor[gray]{0.9}}c}{--}  
            & \multicolumn{1}{>{\cellcolor[gray]{0.9}}c}{--}  
            & \multicolumn{1}{>{\cellcolor[gray]{0.9}}c}{--}  
            & \multicolumn{1}{>{\cellcolor[gray]{0.9}}c}{--} \\
            &Baseline & 66.34 & 54.33  && 44.76   & 30.14 && 8.23 & 4.10 & 47.63 & 52.40 & 49.89\\
            & \textbf{NDPP} (Ours)  & \textbf{67.53} & \textbf{54.93} &&\textbf{46.48}&\textbf{31.06}&&\underline{8.64}&\underline{4.45} & \textbf{48.41}& \textbf{53.53}& \textbf{50.83}\\     
            & \textbf{ODPP} (Ours)  & \underline{66.52 }& \underline{54.85} &&\underline{45.49}&\underline{30.40}&&\textbf{8.72}&\textbf{4.50} & \underline{48.19}& \underline{53.28}& \underline{50.60}\\
\bottomrule
\end{tabular}}
\label{Tab:gldv2_no_pretrain}
\end{table*}

\begin{figure*}[t]
\centering
\includegraphics[width=0.75\textwidth]{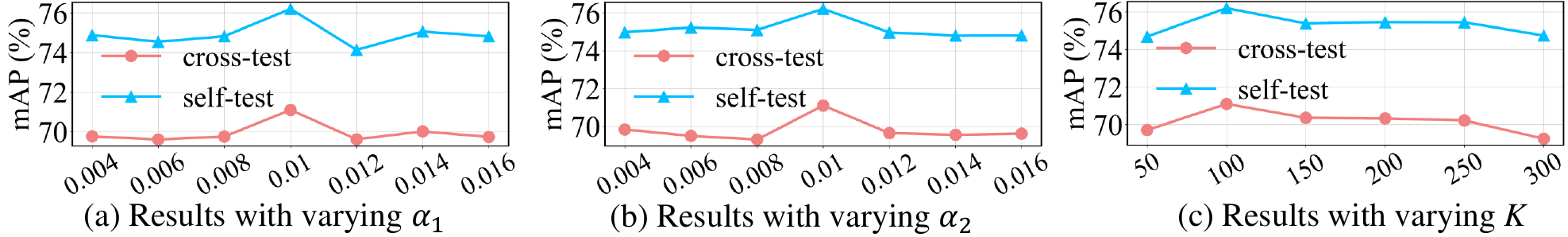}
\vspace{-2mm}
\caption{Experimental results of our proposed NDPP with varying scale factor ($\alpha_1$ and $\alpha_2$) and varying neighbor number ($K$) on the RParis dataset.}
\label{Fig:hyperparameter_ndpp}
\end{figure*}

\begin{figure*}[t]
\centering
\includegraphics[width=0.75\textwidth]{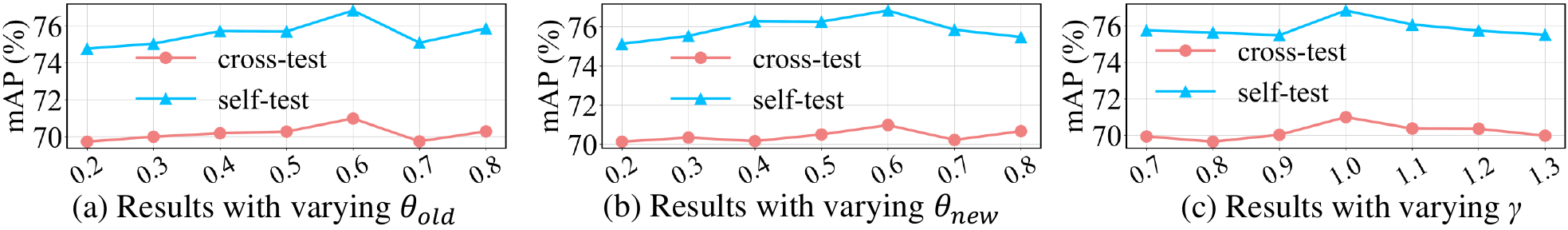}
\vspace{-2mm}
\caption{Experimental results of our proposed ODPP with varying threshold ($\theta_{old}$ and $\theta_{new}$) and varying loss weight ($\gamma$) on the RParis dataset.}
\label{Fig:hyperparameter_odpp}
\vspace{-3mm}
\end{figure*}

\begin{figure*}[t]
\centering
\includegraphics[width=0.9\textwidth]{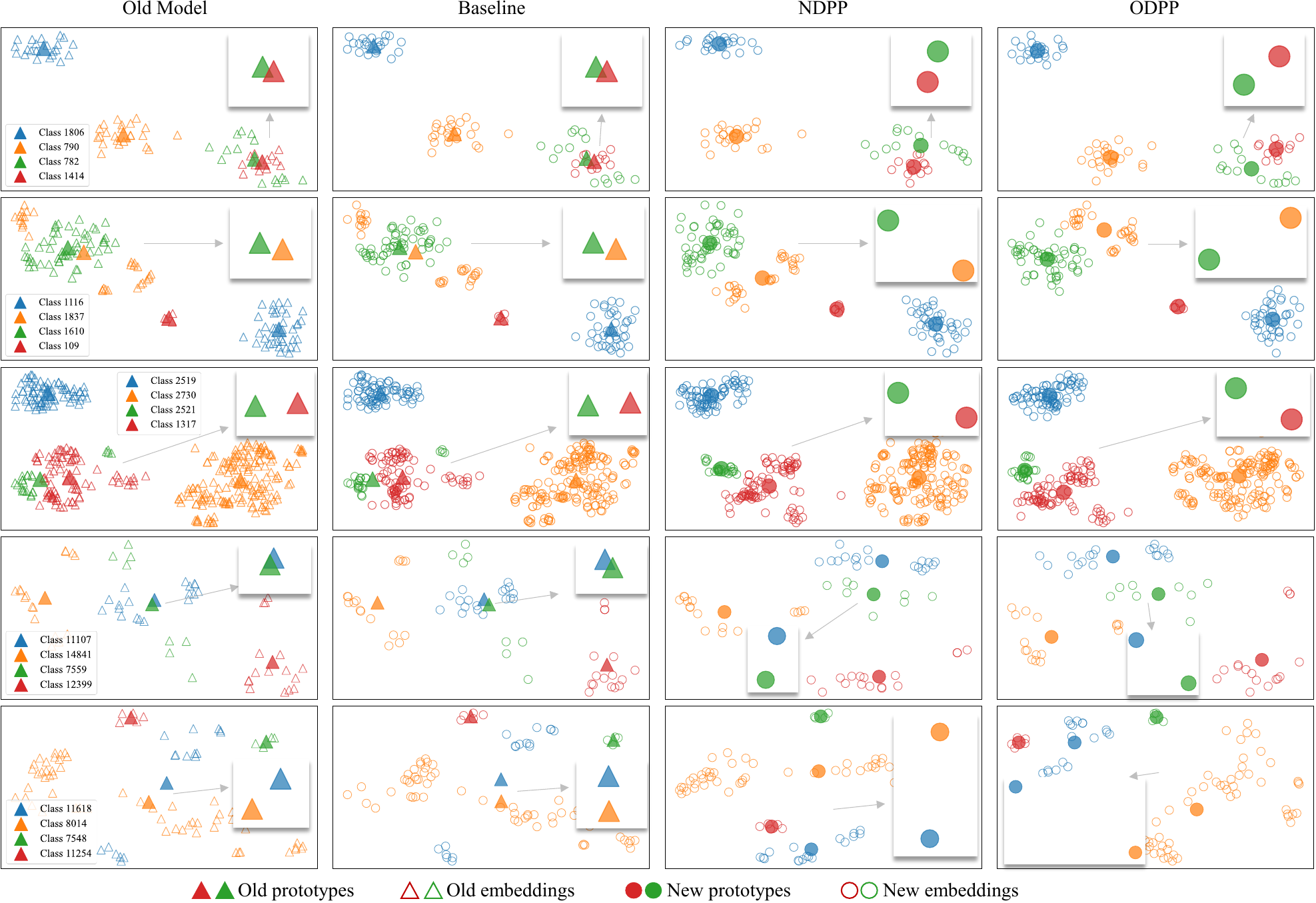}
\vspace{-2mm}
\caption{Distribution of the embeddings and prototypes. Compared to the Baseline, NDPP and ODPP generate more distinct embeddings. For instance, in the first row, NDPP and ODPP produce embeddings that are more distinctly separated for classes 782 and 1414 than those produced by the Baseline. Class IDs below 7318 belong to the old training set, whereas those exceeding 7318 are newly added to the training set.}
\label{Fig:feature_distribution}
\vspace{-3mm}
\end{figure*}

\subsection{Ablation studies}
To analyze our prototype perturbation methods, we conduct ablation studies for single-step backward-compatible learning on GLDv2 using the data-extension setting (9\%$\rightarrow$30\%).

\vspace{2mm}
\noindent\textbf{Analyses on different perturbation methods.} We analyze the effect of different perturbation methods with five variants: 1) Baseline, which performs prototype-based contrastive learning without perturbation; 2) NDPP-old, perturbing the prototypes with only old neighbors; 3) NDPP, perturbing the prototypes based on joint neighbors; 4) ODPP-old, which learns the perturbation only based on old prototypes; 5) ODPP, which learns the perturbation based on joint prototypes. Table~\ref{Tab:ablation} reports the results of the variants. 

Compared with the baseline, both NDPP-old and ODPP-old improve the self-test and cross-test mAP; this demonstrates that our prototype perturbation mechanism can not only enhance the discriminative ability of the new model but also improve its backward compatibility. Compared with NDPP-old and ODPP-old, NDPP and ODPP further improve the self-test mAP and cross-test mAP, respectively. The performance gaps manifest that the leverage of the new feature distribution benefits to acquiring more effective perturbations.

\vspace{2mm}
\noindent\textbf{Effect of the pretrained parameters.} To further analyze the effect of the pretrained parameters, we conduct the BCL experiment using the ResNet18 with randomly initialized parameters. Table~\ref{Tab:gldv2_no_pretrain} presents the results of Baseline, NDPP, and ODPP with and without pretrained parameters. Compared with using pretrained parameters, both the new and old models trained independently demonstrate a significant performance decline when initialized randomly. This indicates the importance of the pretrained parameters in visual retrieval. Notably, the cross-test performance of NDPP and ODPP exceeds the self-test performance of the old model on all three datasets, which highlights the effectiveness of prototype perturbation even in the absence of pretrained parameters. Nevertheless, it remains evident that the overall performance $\mathcal{P}_{1-score}$ without pretrained parameters is inferior to that with pretrained parameters, suggesting that randomly initialized feature spaces hinder the computation of effective prototype perturbations.

\vspace{2mm}
\noindent\textbf{Studies on hyperparameters}. 
We also conduct studies on the hyperparameters through greedy search instead of grid search. For NDPP, we analyze the influence of the scale factors $\alpha_1$ and $\alpha_2$ as well as the neighbor number $K$. Figure~\ref{Fig:hyperparameter_ndpp} presents the performance of NDPP with varying hyperparameters. The model performance initially increases as the two scale factors increase and reaches the peak at both 0.01. Further increasing the scale value may introduce excessive perturbations, degrading the performance. Besides, NDPP achieves the best result at $K=100$. 
For ODPP, we analyze the influence of the thresholds $\theta_{old}$ and $\theta_{new}$ as well as the balance weight $\gamma$. Figure~\ref{Fig:hyperparameter_odpp} illustrates the performance of ODPP with varying hyperparameters. We can observe that the performance improves with increasing the two thresholds and saturates at about 0.6. It implies that imposing a perturbation to the prototypes less similar to others (using a lower threshold) damages the retrieval performance. In addition, ODPP obtains the best result at $\gamma\!=\!1.0$. The experimental results suggest that the importance of the old and new feature spaces in obtaining appropriate perturbations is likely to be nearly identical. Please note that our study on hyperparameters is not to obtain the best performance through hyperparameter fine-tuning. Instead, it is to demonstrate the feasibility of finding a suitable configuration to generate appropriate perturbations that can enhance the discriminative ability of the new model without compromising its compatibility.

Notably, self-test and cross-test performance typically exhibit a positive correlation in the overall trends when varying hyperparameters. The reason is that the hyperparameters jointly determine the perturbations applied to the old prototypes, which further affect the feature distribution of the new model. A well-distributed feature distribution of the new model facilitates improvements in both self-test and cross-test performance. Hence, self-test and cross-test seem correlated. Similar phenomena have also been observed in recent publications~\cite{advbct, AFF}. Besides the feature distribution of the new model, the cross-test performance is also affected by the alignment between the new and old models. Therefore, self-test and cross-test performance sometimes demonstrate divergence in local trends.

\subsection{Analysis on training efficiency}
In addition to retrieval performance, we also evaluate the training efficiency of NDPP and ODPP on GLDv2. Herein, the evaluation experiment is conducted on two Nvidia RTX 3090 GPUs with 30\% of GLDv2 training data. The computation complexity for perturbation calculation based on the new and old prototypes is $3.05\times 10^3$G FLOPs for NDPP and $4.98\times 10^4$G FLOPs for ODPP in a single training epoch. The model training time per epoch is approximately 672.35 seconds for NDPP and 869.75 seconds for ODPP. ODPP increases the training time by approximately 29.36\% compared to NDPP.

\begin{figure}[t!]
\centering
\includegraphics[width=0.9\linewidth]{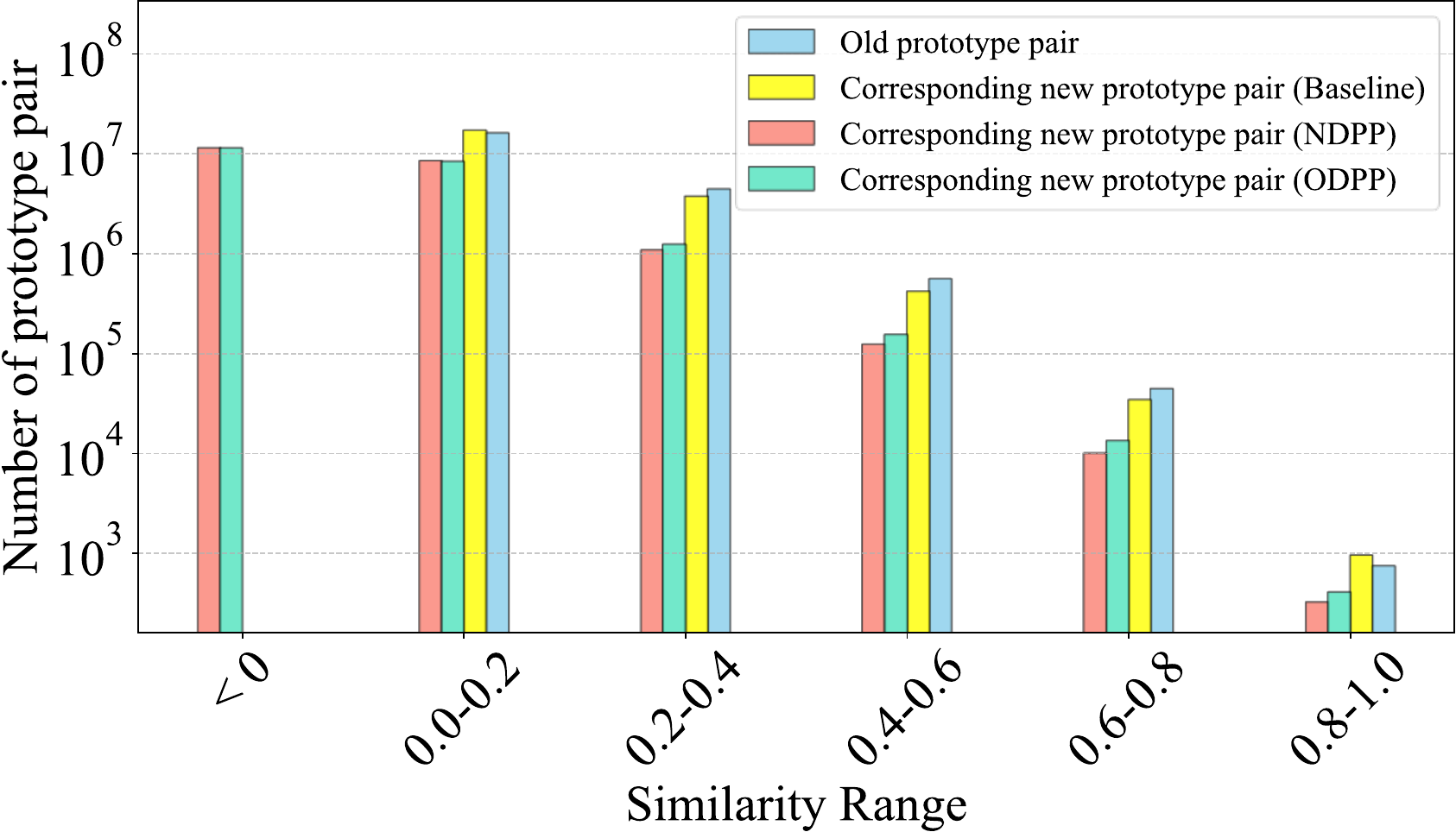}
\vspace{-2mm}
\caption{Distribution of the similarity of the prototype pair. Each bar denotes the number of the new/old prototype pair within a similarity range. Here, we statistically analyze how NDPP and ODPP influence prototype pairs that exhibit \textit{positive cosine similarity} in the feature spaces of the old model and the new one trained with Baseline. This visualization reveals a pronounced decreasing trend in the similarity of these prototype pairs in the new feature space of NDPP and ODPP. Even the similarity of many prototype pairs becomes negative in the new feature space of NDPP and ODPP.}
\label{Fig:pairs_sim}
\vspace{-4mm}
\end{figure}

\subsection{Analysis on feature distribution}
To obtain more insights into the pros and cons of our methods, we first visualize the distribution of the embeddings and prototypes with t-SNE in Figure~\ref{Fig:feature_distribution}. The new embeddings extracted by the Baseline (w/o perturbations) cluster around the nearly indistinguishable old prototypes (\eg~classes 1837 \&~1610 in the first row). As a result, it becomes difficult to differentiate the new embeddings belonging to these classes. The discriminative ability of the new model is degenerated due to the strict backward-compatible constraints. By contrast, the new embeddings extracted by NDPP and ODPP of these classes are distributed more separately, which demonstrates the effectiveness of our prototype perturbation mechanism.

To further assess the effectiveness of our method on similar prototypes, we visualize the distribution of the similarity of the prototype pairs in the embedding space, as illustrated in Figure~\ref{Fig:pairs_sim}. Specifically, we statistically analyze how NDPP and ODPP influence the prototype pairs that have \textit{positive cosine similarity} in both feature spaces of the old model and the new one trained with the baseline method. Each bar in Figure~\ref{Fig:pairs_sim} indicates the number of the new/old prototype pairs within a similarity range. We can observe a pronounced decreasing trend in the similarity of these prototype pairs in the new feature space of NDPP and ODPP. This demonstrates that our method effectively enhances the separability of the prototype pairs that are similar in the old feature space.

Prototype perturbation aims to enhance the uniformity of feature distribution of the new model by providing well-distributed pseudo-old prototypes. This objective shares some similarities with Neural Collapse (NC)~\cite{papyan2020prevalence,zhu2021geometric}, a phenomenon where last-layer features and classifiers exhibit simplified and symmetric geometric structures during the terminal phase of training. Although our settings do not align closely with existing studies~\cite{papyan2020prevalence,zhu2021geometric} on NC, recent works~\cite{wang2020understanding,chen2021intriguing} demonstrated that features learned through contrastive loss, a variant of cross-entropy loss, exhibit properties analogous to those of NC. Specifically, samples from the same class are mapped to nearby features (\textit{alignment}), while features from different classes are maximally separated in feature space (\textit{uniformity}). In Eq.~\eqref{eq:total_loss}, $\mathcal{L}_{ce}$ inherently encourages the new features of different classes to be as separated as possible, while $\mathcal{L}_{bc}$ may negatively affect the uniformity of the new model, when there are indistinguishable classes for the old model. Prototype perturbation can alleviate this issue and enhance the uniformity of the new feature space. We conduct experiments to validate the above analysis. Following~\cite{papyan2020prevalence}, we measure the standard deviation of the cosines between pairs of new prototypes of different methods, including Baseline, NDPP, and ODPP. As shown in Table~\ref{Tab:std}, prototype perturbation contributes to a smaller standard deviation, \ie~benefits to the uniformity of the new feature space.

\begin{table}[t!]
\centering
\setlength{\tabcolsep}{10pt}
\renewcommand{\arraystretch}{1.1}
\caption{The standard deviation of the cosines between pairs of new prototypes of different methods. This experiment is conducted on In-shop using ResNet18 as the backbone.}
\resizebox{0.95\linewidth}{!}{
\begin{tabular}{lcccc}
\toprule
Methods &  $\dagger$R18 (new) &Baseline& NDPP (Ours) & ODPP (Ours)\\
\midrule
Standard deviation $\downarrow$ & 0.0953& 0.1039 & 0.0919 & 0.0890 \\
\bottomrule
\end{tabular}}
\label{Tab:std}
\vspace{-3mm}
\end{table}

\section{Conclusion}
In this paper, we present a prototype perturbation mechanism for BCL. It allows us to align the new feature space with a pseudo-old feature space defined by the perturbed prototypes, thereby preserving the discriminative ability of the new model during BCL. Specifically, we have developed two approaches, neighbor-driven prototype perturbation and optimization-driven prototype perturbation, to implement this mechanism. We have conducted extensive experiments on the landmark and commodity datasets. Compared with state-of-the-art algorithms, our two approaches can achieve superior performance on both self-test and cross-test in most cases, demonstrating their effectiveness.

\bibliographystyle{IEEEtran}
\bibliography{prototype_perturbation}

\begin{IEEEbiography}[{\includegraphics[width=1in,height=1.25in,clip,keepaspectratio]{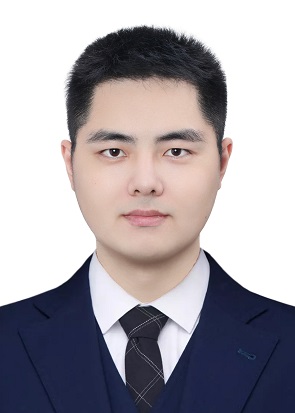}}]{Zikun Zhou}
received his Ph.D. degree from Harbin Institute of Technology in 2022. He is currently a Postdoctoral Research Fellow with Harbin Institute of Technology, Shenzhen. Before joining Harbin Institute of Technology, Shenzhen, He was an assistant research fellow with Pengcheng Laboratory. His research interests include computer vision, LLM agents, and machine learning.
\end{IEEEbiography}
\vspace{-0.25in}
\begin{IEEEbiography}[{\includegraphics[width=1in,height=1.25in,clip,keepaspectratio]{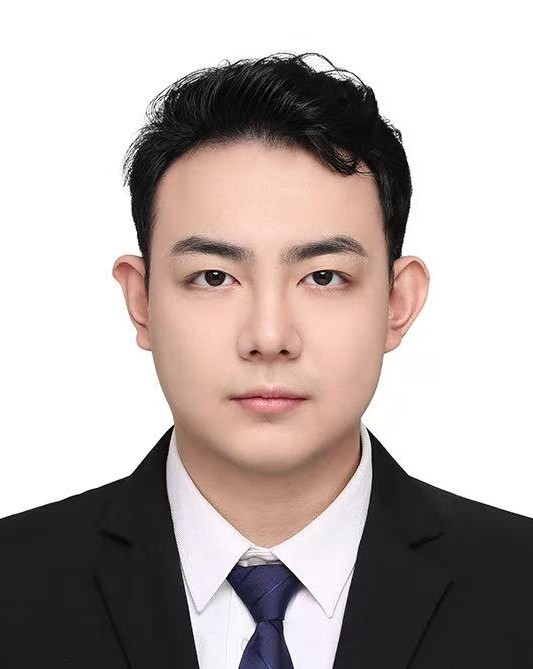}}]{Yushuai Sun} received his B.S. degree from Harbin Institute of Technology in 2023. He is currently pursuing his Ph.D in Harbin Institute of Technology, Shenzhen. His research interests include computer vision and model pruning.
\end{IEEEbiography}
\vspace{-0.25in}
\begin{IEEEbiography}[{\includegraphics[width=1in,height=1.25in,clip,keepaspectratio]{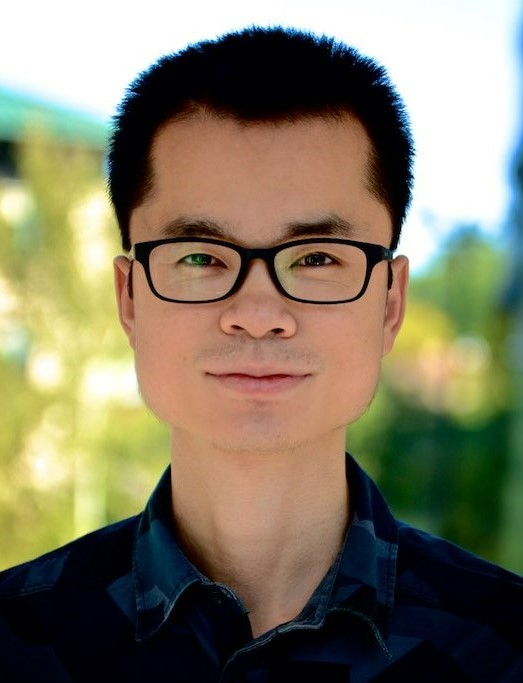}}]{Wenjie Pei} received the Ph.D. degree from the Delft University of Technology, Delft, The Netherlands, working with Dr. Laurens van der Maaten and Dr. David Tax, in 2018. In 2016, he was a Visiting Scholar with Carnegie Mellon University, Pittsburgh, PA, USA. He is currently a Professor with the Harbin Institute of Technology, Shenzhen, China. Before joining Harbin Institute of Technology, he was a Senior Researcher on computer vision with Tencent Youtu X-Laboratory, Shenzhen. His research interests include computer vision and machine learning.
\end{IEEEbiography}
\vspace{-0.25in}
\begin{IEEEbiography}[{\includegraphics[width=1in,height=1.25in,clip,keepaspectratio]{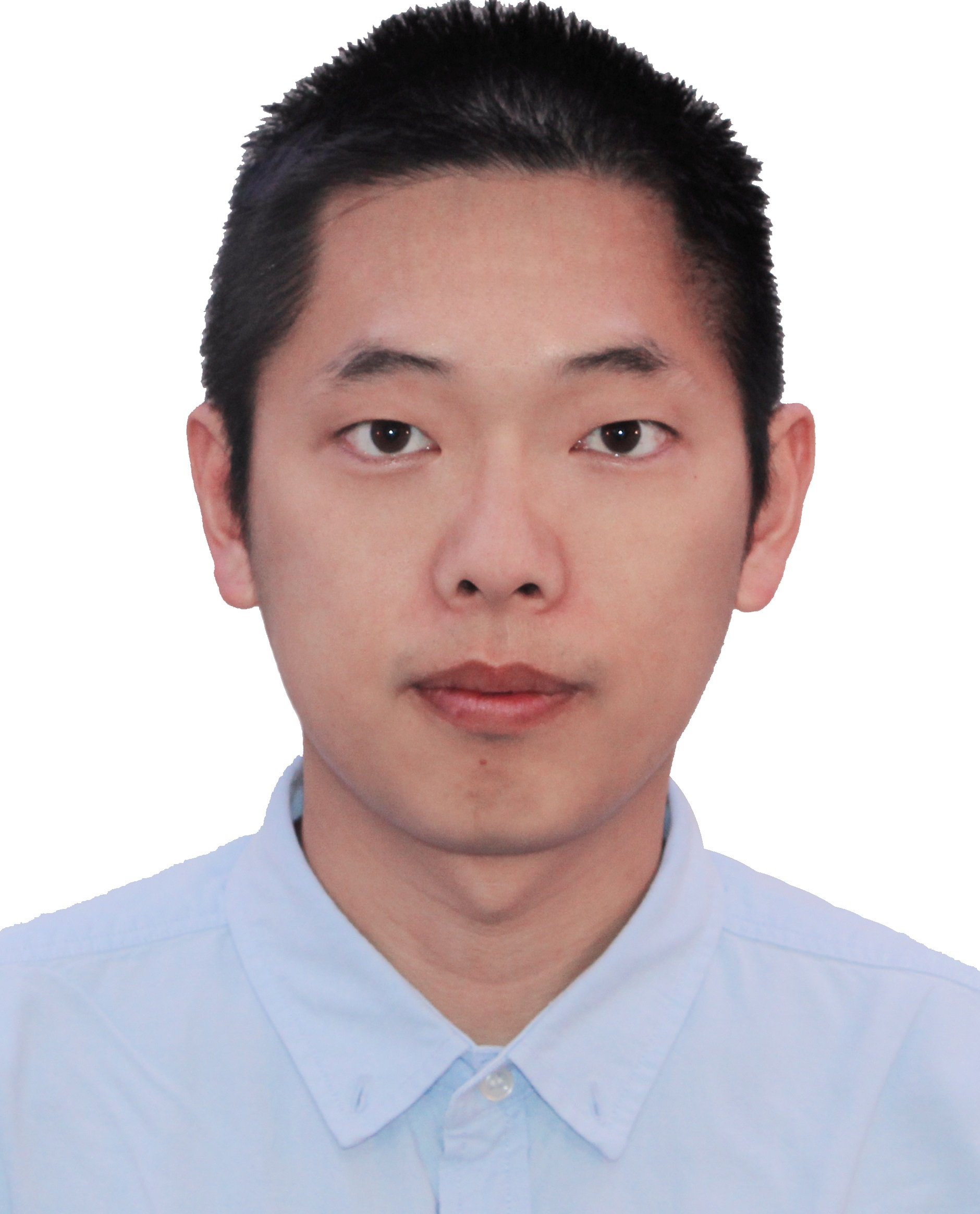}}]{Xin Li}
is currently an associate research fellow with Pengcheng Laboratory, Shenzhen, China. He received the Ph.D. degree computer applied technology from Harbin Institute of Technology, Shenzhen, China, in 2020. He was sponsored by China Scholarship Council as a Visiting Ph.D. Student with the University of California at Merced from 2017 to 2018. His research interests include computer vision and deep learning. 
\end{IEEEbiography}
\vspace{-0.25in}
\begin{IEEEbiography}[{\includegraphics[width=1in,height=1.25in,clip,keepaspectratio]{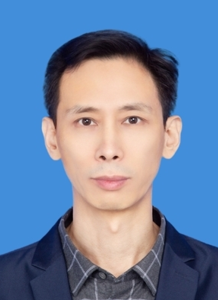}}]{Yaowei Wang} (Member, IEEE) is currently a professor with the Pengcheng Laboratory and Harbin Institute of Technology, Shenzhen, China. He enjoys special government allowances of the State Council. He is the author or co-author of more than 140 technical articles in international journals and conferences, including TOMM, ACM MM, IEEE TIP, CVPR, ICCV, and IJCAI. His current research interests include multimedia content analysis and understanding, machine learning, and computer vision. He serves as the chair of the IEEE Digital Retina Systems Working Group and a member of IEEE, CIE, CCF, CSIG. He was the recipient of the second prize of the National Technology Invention in 2017, the first prize of the CIE Technology Invention in 2015, and the first prize of the CIE Scientific and Technological Progress in 2022.
\end{IEEEbiography}

\end{document}